\documentclass[10pt,twocolumn,letterpaper]{article}

\usepackage{cvpr}
\usepackage{times}
\usepackage{epsfig}
\usepackage[english]{babel}
\usepackage{blindtext}
\usepackage{graphicx,subfigure}
\usepackage{booktabs}
\usepackage{enumerate}
\usepackage{comment}
\usepackage{amsmath}
\usepackage{amssymb}
\usepackage{float}
\usepackage{wrapfig}
\usepackage[normalem]{ulem}
\usepackage{url}
\usepackage[font={footnotesize},labelfont={footnotesize}]{caption}

\newcommand{\detect}{\texttt{detect}\xspace}
\newcommand{\sub}{\texttt{aso-sub}\xspace}
\newcommand{\seq}{\texttt{seq-sub}\xspace}

\newcommand{\glance}{\texttt{glance}\xspace}

\newcommand{\pascal}{PASCAL\xspace}
\newcommand{\ens}{\texttt{ens}\xspace}
\newcommand{\reffig}[1]{Fig.~\ref{#1}}
\newcommand{\refsec}[1]{Sec.~\ref{#1}}

\usepackage[pagebackref=true,breaklinks=true,letterpaper=true,colorlinks,bookmarks=false]{hyperref}
\cvprfinalcopy 
\def\cvprPaperID{386} 

\ifcvprfinal\fi
\begin{document}

\title{Counting Everyday Objects in Everyday Scenes}

\author{
    Prithvijit Chattopadhyay\thanks{\,\,\,Denotes equal contribution.}$\,\,^{,1}$ \quad Ramakrishna Vedantam$^{*,1}$ \quad Ramprasaath R. Selvaraju$^{1}$ \\ \quad Dhruv Batra$^2$ \quad Devi Parikh$^2$\\
    $^1$Virginia Tech \quad $^2$Georgia Institute of Technology\\
    {\tt\small $^{1}$\{prithv1,vrama91,ram21\}@vt.edu} \quad {\tt\small $^{2}$\{dbatra,parikh\}@gatech.edu}\\
}


\maketitle

\begin{abstract}
We are interested in counting the number of instances of object classes in natural, \emph{everyday} images. 
Previous counting approaches tackle the problem in restricted domains such as counting pedestrians in surveillance videos.
Counts can also be estimated from outputs of other vision tasks like object detection. In this work, we build dedicated models for counting designed to tackle the large variance in counts, appearances, and scales of objects found in natural scenes. Our approach is inspired by the phenomenon of subitizing -- the ability of humans to make quick assessments of counts given a perceptual signal, for small count values. Given a natural scene, we employ a divide and conquer strategy while incorporating context across the scene to adapt the subitizing idea to counting. Our approach offers consistent improvements over numerous baseline approaches for counting 
on the PASCAL VOC 2007 and COCO datasets. Subsequently, we study 
how counting can be used to improve object detection. We then show a proof of concept application of our counting methods to the task of Visual Question Answering, by studying the `how many?' questions in the VQA and COCO-QA datasets.
\end{abstract}
\section{Introduction}\label{sec:intro}
\begin{figure}[t]
\includegraphics[width=\columnwidth]{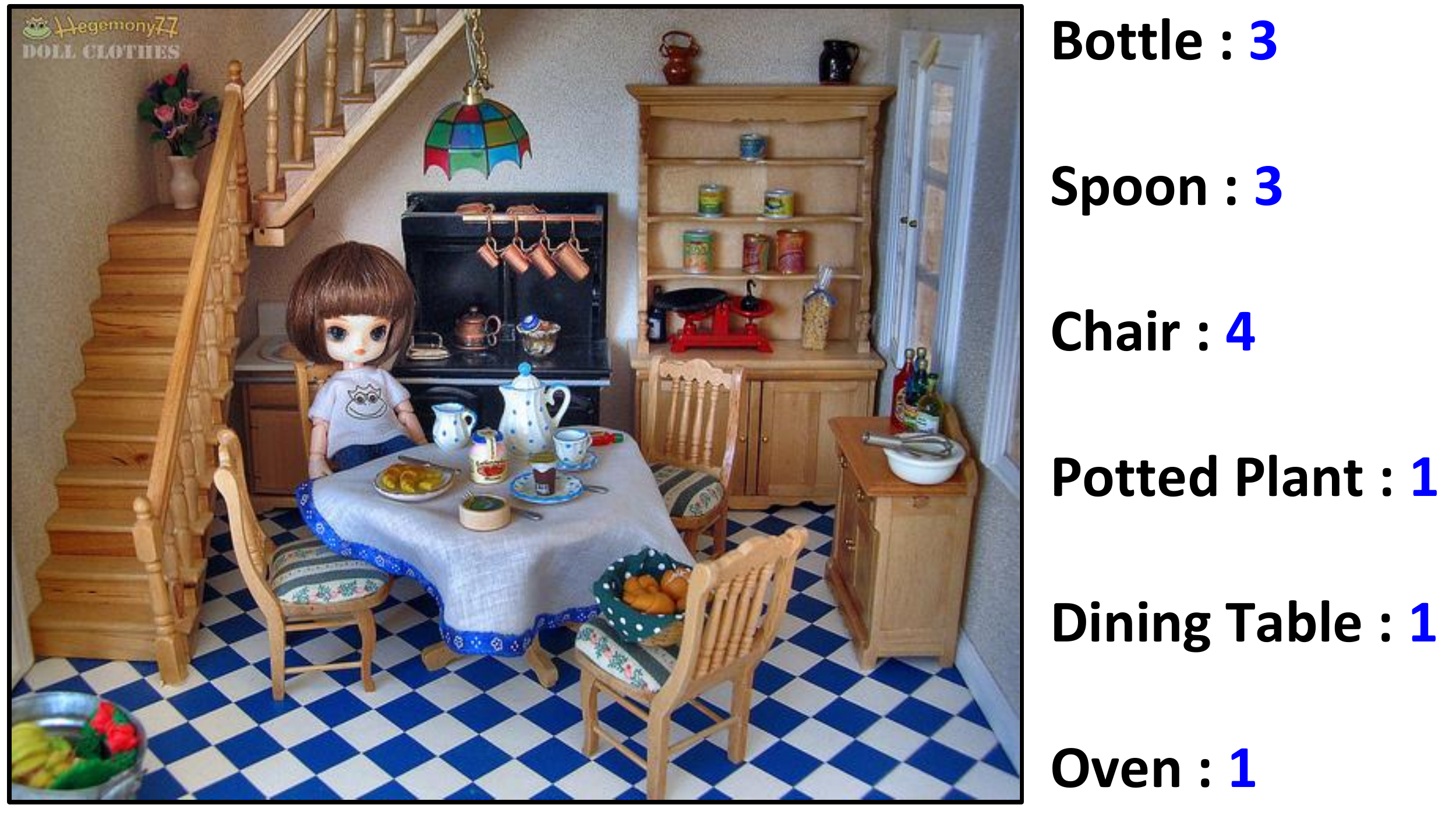}
\vspace{-15pt}
\caption{\footnotesize{We study the problem of counting everyday objects in everyday scenes. Given an everyday scene, we want to predict the number of instances of common objects like bottle, chair \etc.}}
\label{fig:1}
\vspace{-20pt}
\end{figure}

We study the scene understanding problem of counting common objects in natural scenes.
That is, given for example the image in Fig.~\ref{fig:1}, we want to count the number of everyday object categories present in it: for example 4 \emph{chairs}, 1 \emph{oven}, 1 \emph{dining table}, 1 \emph{potted plant} and 3 \emph{spoons}. Such an ability to count seems innate in humans (and even in some animals~\cite{CutiniS}). Thus, as a stepping stone towards Artificial Intelligence (AI), it is desirable to have intelligent machines that can count. 

Similar to scene understanding tasks such as object detection~\cite{Viola2001,Felzenszwalb2008,Girshick2015FastR-CNN,Ren2015FasterNetworks,Gidaris2015ObjectModel,Wan2015End-to-EndSuppression,RedmonDGF15,SSD} and segmentation~\cite{Carreira2012,Long2015FullySegmentation,RenZ16} which require a fine-grained understanding of the scene, object counting is a challenging problem that requires us to reason about the number of instances of objects present while tackling scale and appearance variations.


Another closely related vision task is visual question answering (VQA), where the task is to answer free form natural language questions about an image. Interestingly, questions related to the count of a particular object - \emph{How many red cars do you see?} form a significant portion of the questions asked in common visual question answering datasets~\cite{vqa,Ren2015ExploringAnswering}. Moreover, we observe that end-to-end networks~\cite{vqa,Ren2015ExploringAnswering,Malinowski2015AskImages,FukuiPYRDR16} trained for this task do not perform well on such counting questions. This is not surprising, since the objective is often setup to minimize the cross-entropy classification loss for the correct answer to a question, which ignores ordinal structure inherent to counting. In this work we systematically benchmark how well current VQA models do at counting, and study any benefits from dedicated models for counting on a subset of counting questions in VQA datasets in Sec.~\ref{subsec:vqa_exp_paper}.

Counts can also be used as complimentary signals to aid other vision tasks like detection. If we had an estimate of how many objects were present in the image, we could use that information on a per-image basis to detect that many objects. Indeed, we find that our object counting models improve object detection performance.



We first describe some baseline approaches to counting and subsequently build towards our proposed model.

\begin{figure}
\includegraphics[width=\columnwidth]{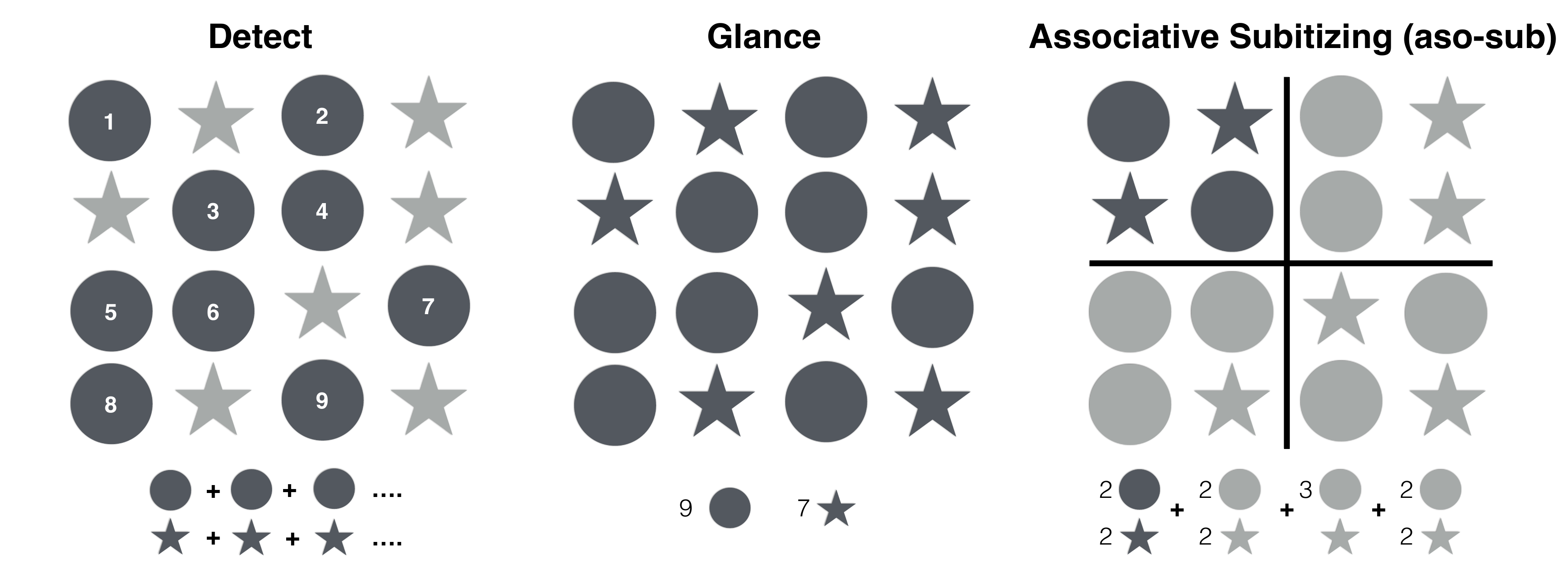}
\vspace{-15pt}
\caption{\footnotesize{A toy example explaining the motivation for three categories of counting approaches explored in this paper. The task is to count the number of stars and circles. In \detect, the idea is to detect instances of a category, and then report the total number of instances detected as the count. In \glance, we make a judgment of count based on a glimpse of the full image. In \sub, we divide the image into regions and judge count based on patterns in local regions. Counts from different regions are added through arithmetic.}}
\label{fig:2}
\vspace{-20pt}
\end{figure}

\noindent \textbf{Counting by Detection:} 
It is easy to realize that perfect detection of objects would imply a perfect count.
While detection is sufficient for counting, localizing objects is not necessary.
Imagine a scene containing a number of mugs kept on a table where the objects occlude each other. In order to count the number of mugs, we need not determine with pixel-accurate segmentations or detections where they are (which is hard in the presence of occlusions) as long as say we can determine the number of handles. Relieving the burden of detecting objects is also effective for counting when objects occur at smaller scales where detection is hard~\cite{Girshick2015FastR-CNN}.
However, counting by detection or \detect still forms a natural approach for counting.

\noindent \textbf{Counting by Glancing:} Representations extracted from Deep Convolutional Neural Networks~\cite{Simonyan2014VeryRecognition,Krizhevsky2012ImageNetNetworks} trained on image classification have been successfully applied to a number of scene understanding tasks such as finegrained recognition~\cite{Donahue2013DeCAF:Recognition}, scene classification~\cite{Donahue2013DeCAF:Recognition}, object detection~\cite{Donahue2013DeCAF:Recognition}, \etc.
We explore how well features from a deep CNN perform at counting through instantiations of our glancing (\glance) models which estimate a global count for the entire scene in a single forward pass. This can be thought of as estimating the count at one shot or \emph{glance}. This is in contrast with \detect, which sequentially increments its count with each detected object (Fig.~\ref{fig:2}). 
Note that unlike detection, which optimizes for a localization objective, the \glance models explicitly \textit{learn} to count.

\noindent \textbf{Counting by Subitizing:} Subitizing is a widely studied phenomenon in developmental psychology~\cite{clements1999subitizing,Klein88,CutiniS} which indicates that children have an ability to directly map a perceptual signal to a numerical estimate, for a small number of objects (typically 1-4). Subitizing is crucial for development and assists arithmetic and reasoning skills. An example of subitizing is how we are able to figure out the number of pips on a face of a die without having to count them or how we are able to reason about tally marks.

Inspired by subitizing, we devise a new counting approach which adopts a divide and conquer strategy, using the additive nature of counts.
Note that \glance can be thought of as an attempt to subitize from a glance of the image. However, as illustrated in Fig.~\ref{fig:2} (center), subitizing is difficult at high counts for humans.

Inspired by this, using the divide and conquer strategy, we divide the image into non-overlapping cells (Fig.~\ref{fig:2} right). We then subitize in each cell and use addition to get the total count. We call this method associative subitizing or \sub.

In practice, to implement this idea on real images, we incorporate context across the cells while sequentially subitizing in each one of them. We call this sequential subitizing or \seq{}. For each of these cells we curate real-valued ground truth, which helps us deal with scale variations.
Interestingly, we found that by incorporating context \seq significantly outperforms the naive subitizing model \sub described above. 
(see Sec.~\ref{sec:counting_results} for more details).

\noindent \textbf{Counting by Ensembling:} It is well known that when humans are given counting problems with large ground truth counts (\eg counting number of pebbles in a jar), individual guesses have high variance, but an average across multiple responses tends to be surprisingly close to the ground truth. This phenomenon is popularly known as the wisdom of the crowd~\cite{Galton}.
Inspired by this, we create an ensemble of counting methods (\ens).

In summary, we evaluate several natural approaches to counting, and propose a novel context and subitizing based counting model. Then we investigate how counting can improve detection. Finally, we study counting questions (`how many?') in the Visual Question Answering (VQA)~\cite{vqa} and COCO-QA~\cite{Ren2015ExploringAnswering} datasets and provide some comparisons with the state-of-the-art VQA models.

\section{Related Work}\label{sec:rel_work}
Counting problems in niche settings have been studied extensively in computer vision~\cite{Zhang2015,Segui2015,Chan2009BayesianCounting,Lempitsky2010}. ~\cite{Chan2009BayesianCounting} explores a Bayesian Poisson regression method on low-level features for counting in crowds. ~\cite{Chan2008PrivacyTracking} segments a surveillance video into components of homogeneous motion and regresses to counts in each region using Gaussian Process regression. Since surveillance scenes tend to be constrained and highly occluded, counting by detection is infeasible. Thus density based approaches are popular. Lempitsky and Zisserman~\cite{Lempitsky2010} count people by estimating object density using low-level features. They show applications on surveillance and cell counting in biological images. Anchovi labs provided users interactive services to count specific objects such as swimming pools in satellite images, cells in biological images, \etc. More recent work constructs CNN-based models for crowd counting~\cite{Zhang2015,onoro2016} and penguin counting~\cite{Arteta16} using lower level convolutional features from shallower CNN models.

Counting problems in constrained settings have a fundamentally different set of challenges to the counting problem we study in this paper. In surveillance, for example, the challenge is to estimate the counts accurately in the presence of large number of ground truth counts, where there might be significant occlusions. In the counting problem on everyday scenes, a larger challenge is the intra-class variance in everyday objects, and high sparsity (most images will have 0 count for most object classes). Thus we need a qualitatively different set of tools to solve this problem.

Other recent work~\cite{zhang2015salient} studies the problem of salient object subitizing (SOS). This is the task of counting the number of salient objects in the image (independent of the category). In contrast, we are interested in counting the number of instances of objects per category. Unlike Zhang~\etal~\cite{zhang2015salient}, who use SOS to improve salient object detection, we propose to improve generic object detection using counts. Our VQA experiments to diagnose counting performance are also similar in spirit to recent work that studies how well models perform on specific question categories (counting, attribute comparison, \etc)~\cite{johnson_CVPR_17} or on compositional generalization~\cite{Agarwal_EMNLP_2016}.

\section{Approach}\label{sec:approach}
Our task is to accurately count the number of instances of different object classes in an image. For training, we use datasets where we have access to object annotations such as object bounding boxes and category wise counts. The count predictions from the models are evaluated using the metrics described in Sec.~\ref{subsec:metrics}.
The input to the \glance, \sub and \seq models are \texttt{fc7} features from a VGG-16~\cite{Simonyan2014VeryRecognition} CNN model. We experiment using both off-the-shelf classification weights from ImageNet~\cite{ILSVRC15} and the detection fine-tuned weights from our \detect models.

\subsection{Detection (\detect{})}\label{app:frcn} We use the Fast R-CNN~\cite{Girshick2015FastR-CNN} object detector to count. Detectors typically perform two post processing steps on a set of preliminary boxes: non maximal suppression (NMS) and score thresholding. NMS discards highly overlapping and likely redundant detections (using a threshold to control the overlap), whereas the score threshold filters out all detections with low scores. 

We steer the detector to count better by varying these two hyperparameters to find the setting where counting error is the least. We pick these parameters using grid search on a held-out val set. For each category, we first select a fixed NMS threshold of 0.3
for all the classes and vary the score threshold between 0 and 1. We then fix the score threshold to the best value and vary the NMS threshold from 0 to 1. 

\subsection{Glancing (\glance{})}\label{app:glance}
Our \glance approach repurposes a generic CNN architecture for counting by training a multi-layered perceptron (MLP) with a L2 loss to regress to image level counts from deep representations extracted from the CNN. The MLP has batch normalization~\cite{Ioffe2015} and Rectified Linear Unit (ReLU) activations between hidden layers. The models were trained with a learning rate of $10^{-3}$ and weight decay set to 0.95. We experiment with choices of a single hidden layer, and two hidden layers for the MLP, as well as the sizes of the hidden units. More details and ablation studies can be found in appendix.

\vspace{-8pt}
\subsection{Subitizing (\sub{}, \seq{})}\label{app:aso-sub}
In our \emph{subitizing} inspired methods, we divide our counting problem into sub-problems on each cell in a non-overlapping grid, and add the predicted counts across the grid. 
\begin{figure}[t]
\includegraphics[width=\columnwidth]{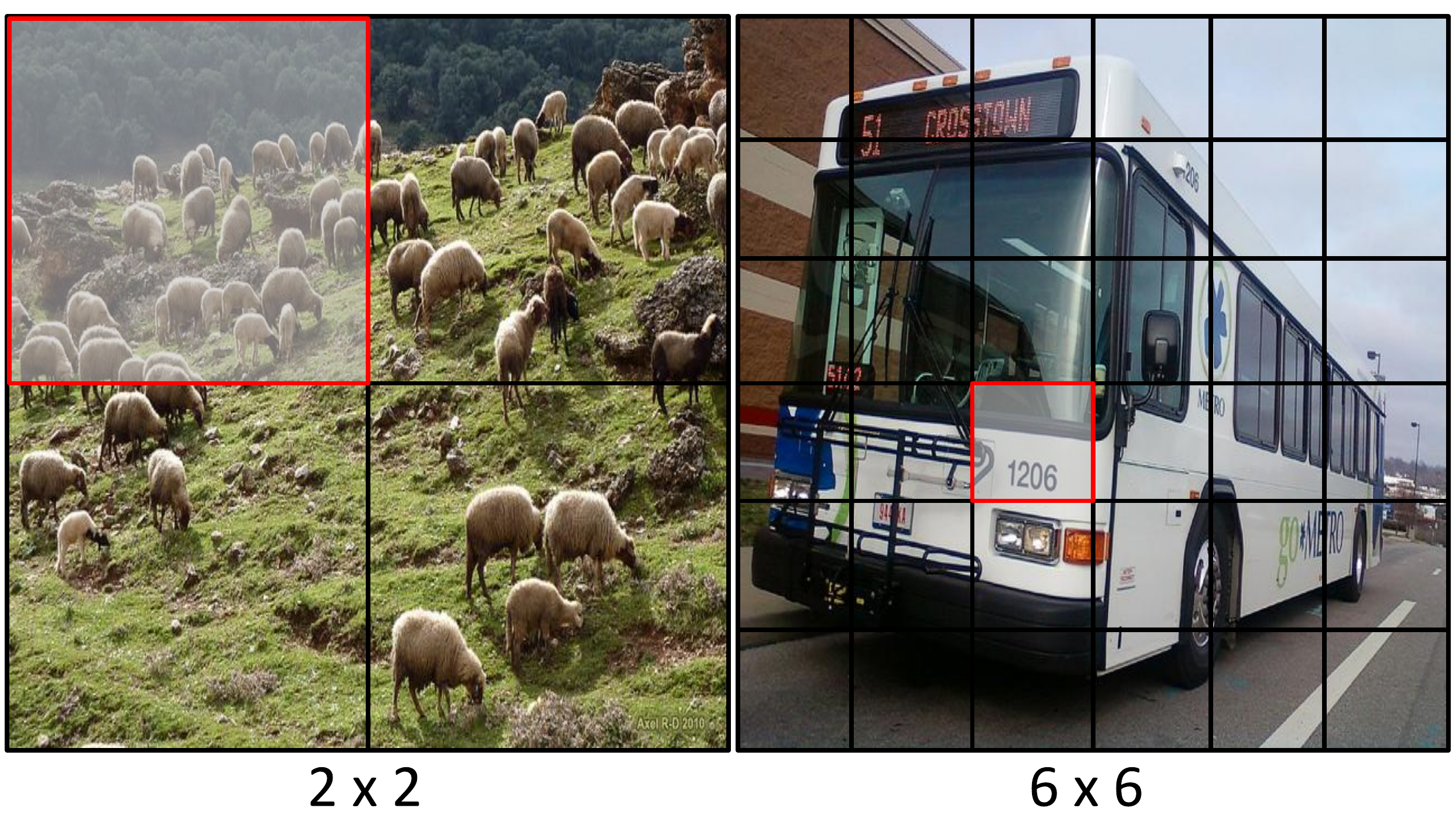}
\vspace{-15pt}
\caption{\footnotesize{\textbf{Canonical counting scale:} Consider images with grids $2\times2$ (left) and $6\times6$ (right). Notice the red cells in both images: it is evident that if the cell size is too large compared to the object (left), it is difficult to estimate the large integer count of `sheep' in the cell. However, if the cell is too small (right), it might be hard to estimate the small fractional count of `bus' in the cell. Hence, we hypothesize that there exists a sweet spot in discretization of the cells that would results in optimum counting performance.}}
\label{fig:cell_example}
\vspace{-20pt}
\end{figure}
In practice, since objects in real images occur at different scales, such cells might contain fractions of an object. We adjust for this by allowing for real valued ground truth. If a cell overlapping an object is very small compared to the object, the small fractional count of the cell might be hard to estimate. On the other hand, if a cell is too large compared to objects present it might be hard to estimate the large integer count of the cell (see Fig.~\ref{fig:cell_example}). This trade-off suggests that at some canonical resolution, we would be able to count the smaller objects more easily by subitizing them, as well as predict the partial counts for larger objects. 
More concretely, we divide the image $I$, into a set of $n$ non-overlapping cells $P = \{p_{1},\cdots,p_{n}\}$ such that $I = \bigcup_{i=1}^np_{i}$ and ${p_{i}\cap{}p_{j}}_{(i\neq j)}=\phi$. Given such a partition $P$ of the image $I$ and associated CNN features $X = \{x_{i},\cdots,x_{n}\}$, we now explain our models based on this approach:
\par \noindent
\textbf{\sub{} :} Our naive \sub{} model treats each cell independently to regress to the real-valued ground truth. We train on an augmented version of the dataset where the dataset size is $n$-fold ($n$ cells per image). Unlike \glance, where feature extracted on the full image is used to regress to integer valued counts, \sub models regress to real-valued counts on non-overlapping cells from features extracted per cell.
Given class instance annotations as bounding boxes  $b = \{b_{1},\cdots,b_{N}\}$ for a category $k$ in an image $I$, we compute the ground truth partial counts ($c_{gt}^k$) for the grid-cells ($p_{i}$) to be used for training as follows: 
\begin{equation}
p_{i}: c_{gt}^k = \sum_{j=1}^{N} \frac{p_{i} \cap b_{j}}{b_{j}} 
\end{equation}
We compute the intersection of each box $b_{i}$ with the cell $p_{i}$ and add up the intersections normalized by $b_{i}$.
Further, given the cell-level count predictions $c_{p_i}$, the image level count prediction is computed as $c=\sum_{i=1}^nmax(0,c_{p_i})$. We use max to filter out negative predictions. 

We experiment with dividing the image into equally sized $3\times3$, $5\times5$, and $7\times7$ grid-cells. The architecture of the models trained on the augmented dataset are the same as \glance. For more details, refer to appendix.
\begin{figure}[t]
\includegraphics[width=\columnwidth]{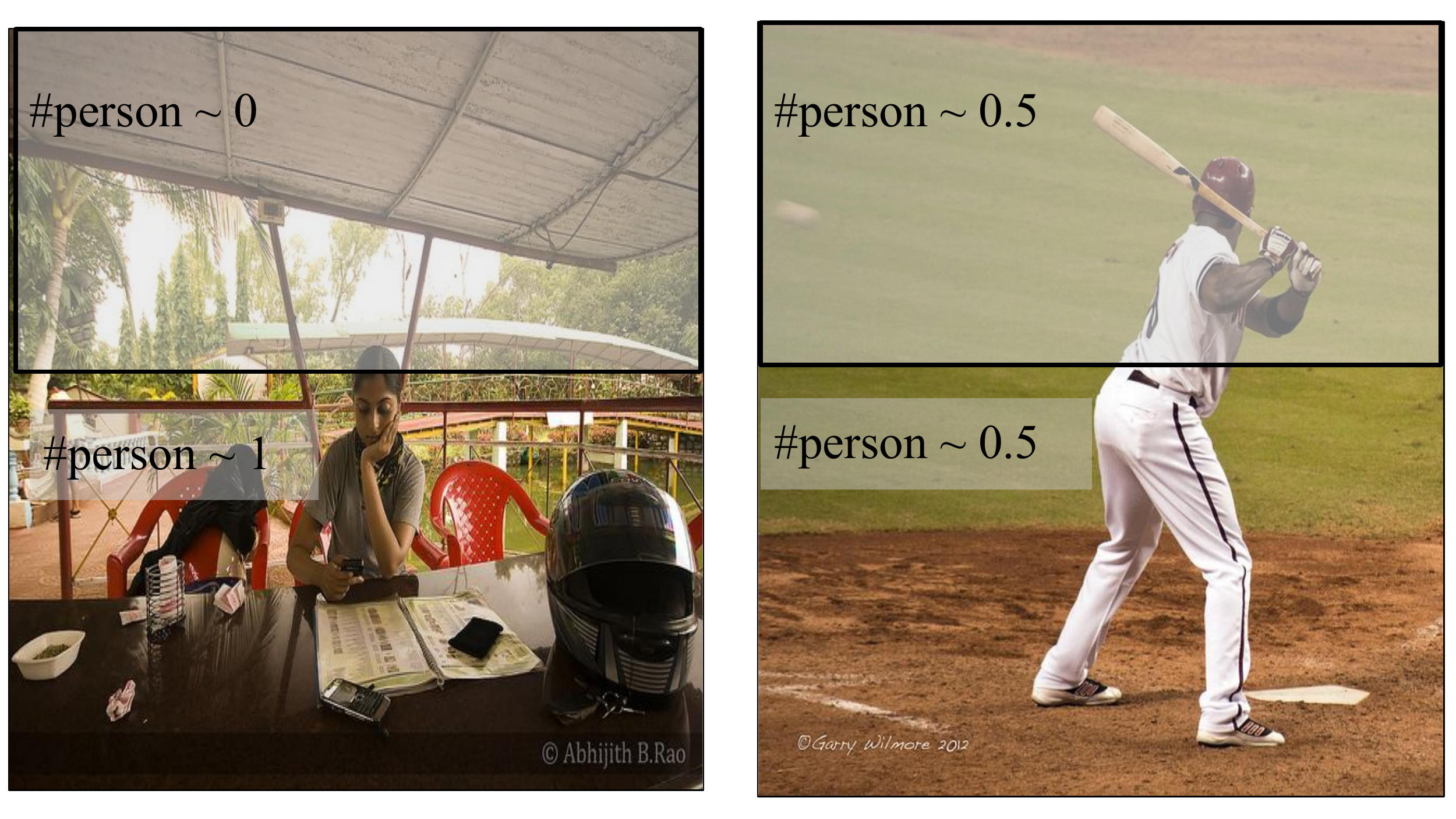}
\vspace{-15pt}
\caption{\footnotesize{For both these images, the count of \emph{person} is $1$. Consider splitting this image into 2 $\times$ 1 cells (for illustration) for \sub. The bottom half of the left image and top half of the right image both contains similar visual signals -- top-half of a person. However, the ground truth count on the cell on the left is 1, and the one on the right is 0.5. An approach that estimates counts from individual cells out of context is bound to fail at these cases. This motivates our proposed approach \seq. 
}}
\label{fig:asoproblem}
\vspace{-17pt}
\end{figure}

\par \noindent
\textbf{\seq{} :} We motivate our proposed \seq{} (Sequential Subitizing) approach by identifying a potential flaw in the naive \sub{} approach. Fig.~\ref{fig:asoproblem} reveals the limitation of the \sub{} model.
If the cells are treated independently, the naive \sub model will be unaware of the partial presence of the concerned object in other cells. This leads to situations where similar visual signals need to be mapped to partial and whole presence of the object in the cells (see Fig.~\ref{fig:asoproblem}). This is especially pathological since Huber or L-2 losses cannot capture this multi-modality in the output space, since the implicit density associated with such losses is either laplacian or gaussian.

Interestingly, a simple solution to mitigate this issue is to model context, which resolves this ambiguity in counts. That is, if we knew about the partial class presence in other cells, we could use that information to predict the correct cell count. Thus, although the independence assumption in \sub is convenient, it ignores the fact that the augmented dataset is not IID. While it is important to reason at a cell level, it is also necessary to be aware of the global image context to produce meaningful predictions. In essence, we propose \seq, that takes the best of both worlds from \glance and \sub.

\par
\begin{figure}
\includegraphics[width=\columnwidth]{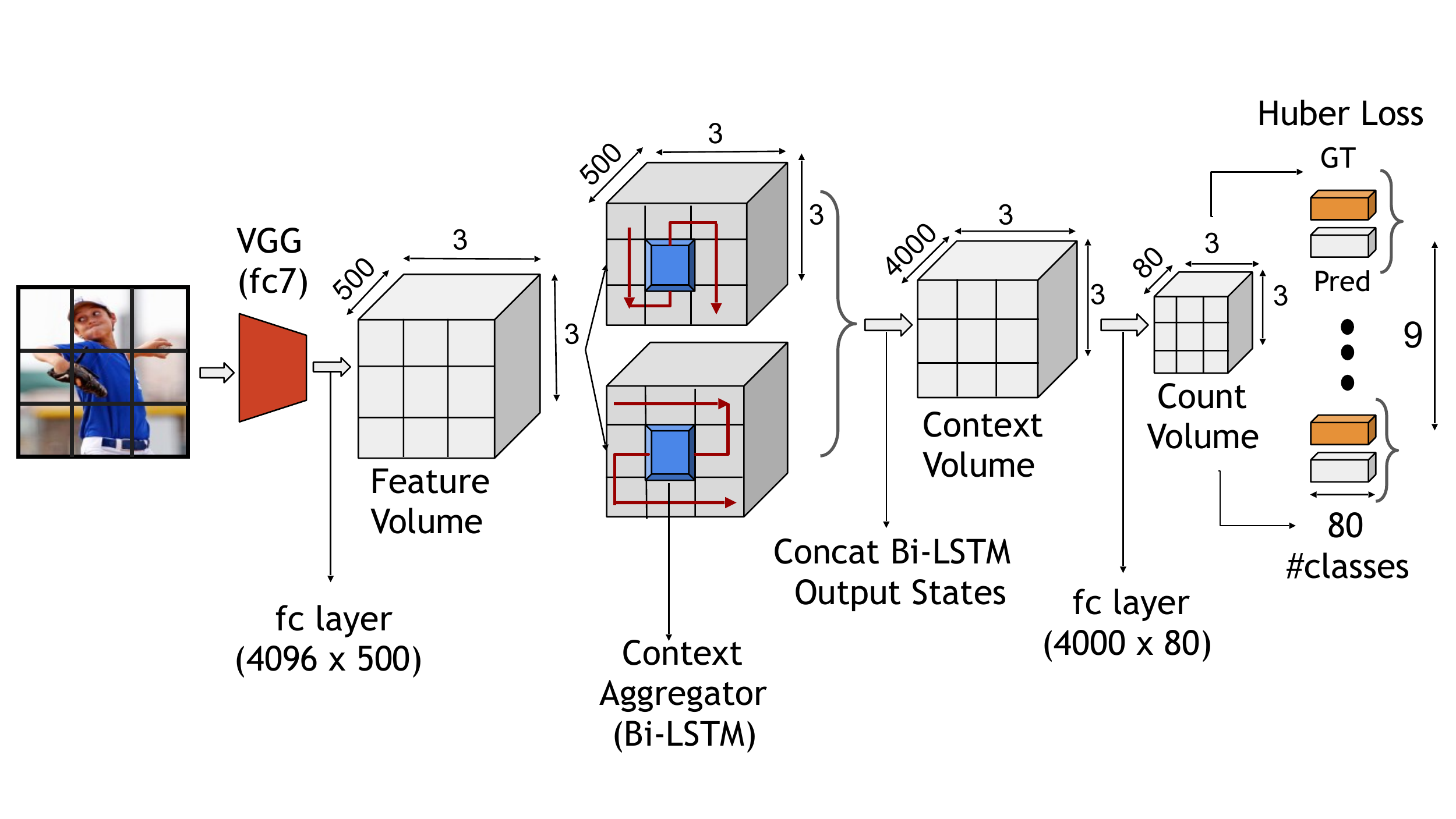}
\vspace{-25pt}
\caption{\footnotesize{Architecture used for our \seq{} models. We extract a hidden layer representation of the \texttt{fc7} feature volume corresponding to the $3\times3$ discretization of the image. Subsequently, we traverse this representation volume in two particular sequences in parallel as shown via two stacked bi-LSTMs per sequence and aggregate context over the image. We get output states corresponding to each of the cells and subsequently get cell-counts via another hidden layer. The hidden layers use ReLU as non-linearity.}}
\label{fig:seqsub}
\vspace{-17pt}
\end{figure}

 \noindent
The architecture of \seq{} is shown in Fig.~\ref{fig:seqsub}. It consists of a pair of 2 stacked bi-directional sequence-to-sequence LSTMs~\cite{Schuster1997BidirectionalRN}. We incorporate context across cells as 
\begin{equation}
c_{p_{i}} = h(f_1(x_1,\theta_1),\cdots,f_n(x_n,\theta_n),i,\theta)
\end{equation}
where individual $f_i(x_i,\theta_i)$ are hidden layer representations of each cell feature with respective parameters and $h(.,\theta)$ is the mechanism that captures context. This can be broken down as follows. Let $H$ be the set containing $f_i(x_i,\theta_i)$s. Let $H_{O1}$ and $H_{O2}$ be 2 ordered sets which are permutations of $H$ based on 2 particular sequence structures. The (traversal) sequences, as we move across the grid in the feature column, is decided on the basis of nearness of cells  (see Fig.~\ref{fig:seqsub}). We experiment with the sequence structures best described for a $3\times3$ grid as \reflectbox{N} and Z which correspond to $H_{O1}$ and $H_{O2}$. Each of these feature sequences are then fed to a pair of stacked Bi-LSTMs ($L_{j}(.,i,\theta_{l})$) and the corresponding cell output states are concatenated to obtain a context vector ($v_{i}$) for each cell as $v_{i} = L_{1}(H_{O1},i,\theta_{l})||L_{2}(H_{O2},i,\theta_{l})$. The cell counts are then obatined as $c_{p_{i}} = g(v_{i},\theta_{g})$. The composition of $L_{j}(.,i,\theta_{l})$ and $g(.,\theta_{g})$ implements $h(.,\theta)$.

We use a Huber Loss objective to regress to the count values with a learning rate of $10^{-4}$ and weight decay set to 0.95. For optimization, we use Adam~\cite{KingmaB14} with a minibatch size of 64. The ground truth construction procedure for training and the count aggregation procedure for evaluation are as defined in \sub.

\section{Experimental Setup}\label{sec:exp_setup}
\subsection{Datasets} 
We experiment with two datasets depicting everyday objects in everyday scenes: the \pascal VOC 2007~\cite{Everingham15} and COCO~\cite{LinECCV14coco}. 
\begin{figure}[t]
\includegraphics[width=\columnwidth]{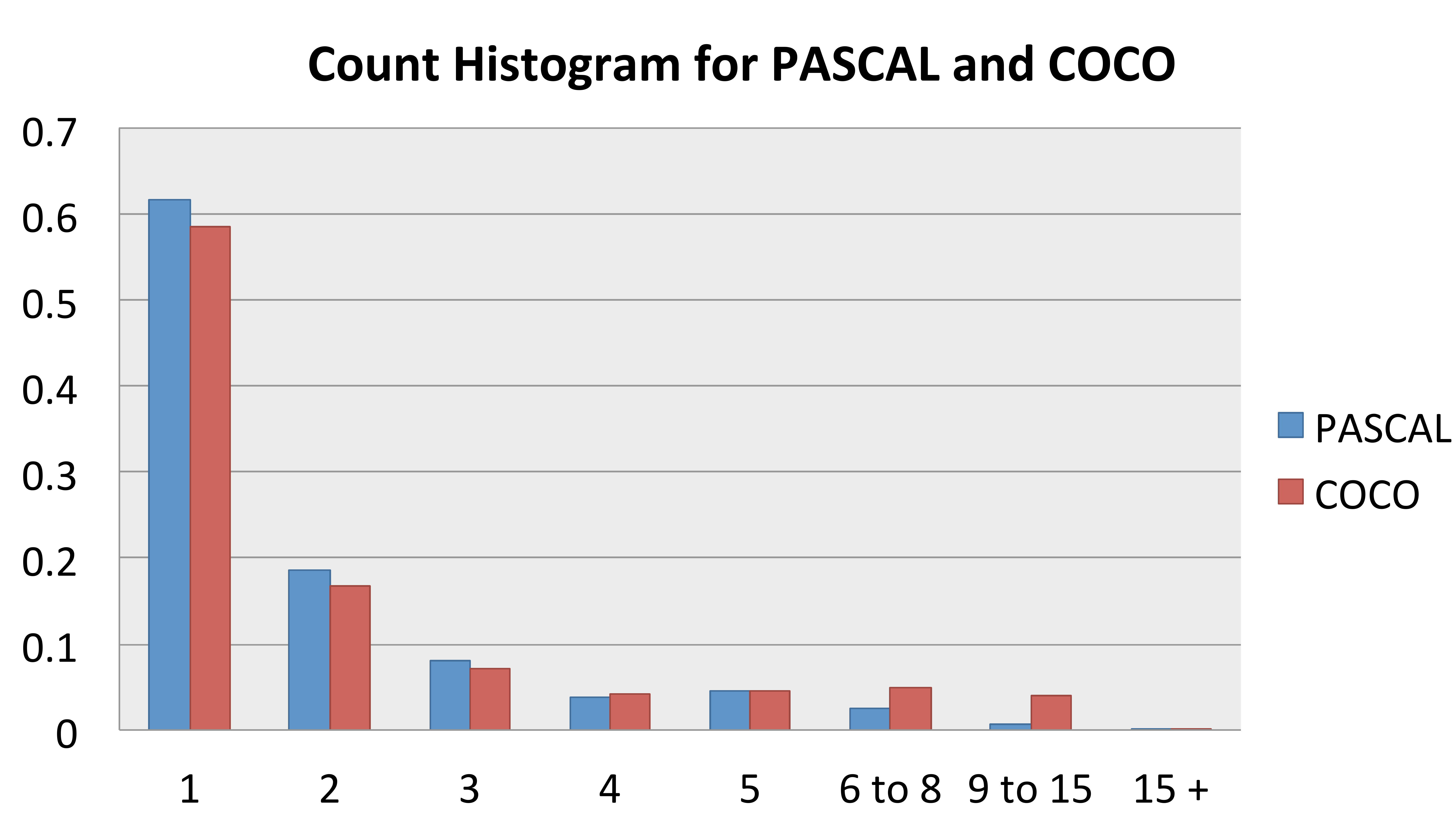}
\vspace{-15pt}
\caption{\footnotesize{Histogram of the non-zero counts in the PASCAL (blue) and COCO (red) datasets across all objects categories and images.}}
\label{fig:histograms}
\vspace{-15pt}
\end{figure}
The PASCAL VOC dataset contains a train set of $2501$ images, val set of $2510$ images and a test set of $4952$ images, and has $20$ object categories. The COCO dataset contains a train set of $82783$ images and a val set of $40,504$ images, with $80$ object categories. On PASCAL, we use the val set as our Count-val set and the test set as our Count-test set. On COCO, we use the first half of val as the Count-val set and the second half of val as the Count-test set. The most frequent count per object category (as one would expect in everyday scenes) is $0$. Fig.~\ref{fig:histograms} shows a histogram of non-zero counts across all object categories. It can be clearly seen that although the two datasets have a fair amount of count variability, there is a clear bias towards lower count values. Note that this is unlike the crowd-counting datasets, in particular~\cite{Idrees} where mean count is $1279.48$ $\pm$ $960.42$ and also unlike PASCAL and COCO, the images have very little scale and appearance variations in terms of objects. 

\subsection{Evaluation}\label{subsec:metrics}
We adopt the root mean squared error (RMSE) as our metric.
We also evaluate on a variant of RMSE that might be better suited to human perception. The intuition behind this metric is as follows. In a real world scenario, humans tend to perceive counts in the logarithmic scale~\cite{dehaene2008log}.
That is, a mistake of 1 for a ground truth count of 2 might seem egregious but the same mistake for a ground truth count of 25 might seem reasonable. Hence we scale each deviation by a function of the ground truth count.

We first post-process the count predictions from each method by thresholding counts at 0, and rounding predictions to closest integers to get predictions $\hat{c_{ik}}$. Given these predictions and ground truth counts $c_{ik}$ for a category $k$ and image $i$, we compute RMSE as follows:
\vspace{-5pt}
\begin{equation}
RMSE_{k} = \sqrt{\frac{1}{N}\sum_{i=1}^{N}(\hat{c_{ik}}-c_{ik})^2}
\end{equation}

and relative RMSE as:
\vspace{-5pt}
\begin{equation}
relRMSE_{k} = \sqrt{\frac{1}{N}\sum_{i=1}^{N}\frac{(\hat{c_{ik}}-c_{ik})^2}{c_{ik}+1}}
\end{equation}

where $N$ is the number of images in the dataset. We then average the error across all categories to report numbers on the dataset (\textbf{mRMSE} and \textbf{m-relRMSE}). 

We also evaluate the above metrics for ground truth instances with non-zero counts. This reflects more clearly how accurate the counts produced by a method (beyond predicting absence) are. 


\subsection{Methods and Baselines}
We compare our approaches to the following baselines:

\noindent \textbf{\texttt{always-0}}: predict most-frequent ground truth count (0).

\noindent \textbf{\texttt{mean}}: predict the average ground truth count on the Count-val set.

\noindent \textbf{\texttt{always-1}}: predict the most frequent non-zero value (1) for all classes.

\noindent \textbf{\texttt{category-mean}}: predict the average count \emph{per} category on Count-val.

\noindent \textbf{\texttt{gt-class}}: treat the \emph{ground truth} counts as classes and predict the counts using a classification model trained with cross-entropy loss.


We evaluate the following variants of counting approaches (see \refsec{sec:approach} for more details):

\noindent \textbf{\detect{}:} We compare two methods for \detect. The first method finds the best NMS and score thresholds as explained in Sec.~\ref{app:frcn}. The second method uses vanilla Fast R-CNN as it comes out of the box, with the default NMS and score thresholds.


\noindent \textbf{\glance{}:} We explore the following choices of features: (1) vanilla classification \texttt{fc7} features \texttt{noft}, (2) detection fine tuned \texttt{fc7} features \texttt{ft}, (3) \texttt{fc7} features from a CNN trained to perform Salient Object Subitizing \texttt{sos}~\cite{zhang2015salient} and (4) flattened \texttt{conv-3} features from a CNN trained for classification  

\noindent \textbf{\sub{}, \seq{}:} We examine three choices of grid sizes (Sec.~\ref{app:aso-sub}): $3\times3$, $5\times5$, and $7\times7$ and \texttt{noft} and \texttt{ft} features as above.  

\noindent \textbf{\ens{}:} We take the best performing subset of methods and average their predictions to perform counting by ensembling (\ens{}).

\section{Results}\label{sec:res}
All the results presented in the paper are averaged on 10 random splits of the test set sampled with replacement. 

\subsection{Counting Results}\label{sec:counting_results}
\par \noindent
\textbf{PASCAL VOC 2007 : }We first present results (Table.~\ref{table:pascal}) for the best performing variants (picked based on the val set) of each method. We see that \seq{} outperforms all other methods. 
Both \glance{} and \detect{} which perform equally well as per both the metrics, while \glance does slightly better on both metrics when evaluated on non-zero ground truth counts. 
To put these numbers in perspective, we find that the difference of $0.01$ $mRMSE$-$nonzero$ between \seq and \sub leads to a difference of 0.19\% mean F-measure performance in our counting to improve detection application (\refsec{subsec:cntdet}). 
We also experiment with \texttt{conv3} features to regress to the counts, similar to Zhang.et.al.~\cite{Zhang2015}. We find that \texttt{conv3} gets $mRMSE$ of 0.63 which is much worse than \texttt{fc7}. We also tried PCA on the \texttt{conv3} features but that did not improve performance. This indicates that our counting task is indeed more high level and needs to reason about objects rather than low-level textures. We also compare our approach with the SOS model~\cite{zhang2015salient} by extracting fc7 features from a model trained to perform category-independent salient object subitizing. We observe that our best performing \glance{} setup using Imagenet trained VGG-16 features outperforms the one using SOS features. This is also intuitive since SOS is a category independent task, while we want to count number of object instances of each category. Finally, we observe that the performance increment from \sub{} to \seq{} is not statistically significant. We hypothesize that this is because of the smaller size of the PASCAL dataset. Note that we get more consistent improvements on COCO (Table.~\ref{table:coco}), which is not only a larger dataset, but also contains scenes that are contextually richer.\footnote{When the Count-val split is considered, PASCAL has an average of $1.98$ annotated objects per scene, unlike COCO which has $7.22$ annotated objects per scene.}
\begin{table}[t] \scriptsize
\setlength{\tabcolsep}{3pt}
\begin{center}
\resizebox{1\columnwidth}{!}{
\begin{tabular}{@{} l  c  c  c  c  c@{}}
\toprule
Approach & mRMSE & mRMSE-nz & m-relRMSE & m-relRMSE-nz\\
\midrule
\texttt{always-0} & 0.66 $\pm$ 0.02 & 1.96 $\pm$ 0.03 & 0.28 $\pm$ 0.03 & 0.59 $\pm$ 0.00\\
\texttt{mean} & 0.65 $\pm$ 0.02 & 1.81 $\pm$ 0.03 & 0.31 $\pm$ 0.01 & 0.52 $\pm$ 0.00\\
\texttt{always-1} & 1.14 $\pm$ 0.01 & 0.96 $\pm$ 0.03 & 0.98 $\pm$ 0.00 & 0.17 $\pm$ 0.03\\
\texttt{category-mean} & 0.64 $\pm$ 0.02 & 1.60 $\pm$ 0.03 & 0.30 $\pm$ 0.00 & 0.45 $\pm$ 0.00\\
\texttt{gt-class} &  0.55 $\pm$ 0.02 & 2.12 $\pm$ 0.07 & 0.24 $\pm$ 0.00 & 0.88 $\pm$ 0.01\\
\detect{}  &  0.50 $\pm$ 0.01 & 1.92 $\pm$ 0.08 & 0.26 $\pm$ 0.01 & 0.85 $\pm$ 0.02\\
\midrule
\glance{}\texttt{-noft-2L} & 0.50 $\pm$ 0.02 & 1.83 $\pm$ 0.09 & 0.27 $\pm$ 0.00 & 0.73 $\pm$ 0.00\\
\glance{}\texttt{-sos-2L} & 0.51 $\pm$ 0.02 & 1.87 $\pm$ 0.08 & 0.29 $\pm$ 0.01 & 0.75 $\pm$ 0.02 \\
\sub{}\texttt{-ft-1L-$3\times3$} & 0.43 $\pm$ 0.01 & 1.65 $\pm$ 0.07 & 0.22 $\pm$ 0.01 & 0.68 $\pm$ 0.02 \\
\seq{}\texttt{-ft-$3\times3$} &\textbf{0.42 $\pm$ 0.01} &\textbf{1.65 $\pm$ 0.07} & 0.21 $\pm$ 0.01 & 0.68 $\pm$ 0.02 \\
\ens{}& 0.42 $\pm$ 0.17 & 1.68 $\pm$ 0.08 &\textbf{0.20 $\pm$ 0.00} &\textbf{0.65 $\pm$ 0.01}\\
\bottomrule
\end{tabular}
}
\vspace{0pt}
\caption{\footnotesize{Counting performance on PASCAL VOC 2007 Count-test Set (\texttt{L} implies the number of hidden layers). \textbf{Lower} is better. \ens is a combination of \glance{}\texttt{-noft-2L}, \sub{}\texttt{-ft-1L-$3\times3$} and \seq{}\texttt{-ft-$3\times3$}.\vspace{4pt}}}
\label{table:pascal}
\vspace{-33pt}
\end{center}
\end{table}
\par \noindent
\textbf{COCO :} We present results for the best performing variants (picked based on the val set) of each method. The results are summarized in Table.~\ref{table:coco}. We find that \seq does the best on both $mRMSE$ and $m$-$relRMSE$ as well as their non-zero variants by a significant margin. A comparison indicates that the \texttt{always-0} baseline does better on COCO than on PASCAL. This is because COCO has many more categories than PASCAL. Thus, the chances of any particular object being present in an image decrease compared to PASCAL. The performance jump from \sub to \seq here is much more compared to \pascal. Recent work by Ren and Zemel~\cite{RenZ16} on Instance Segmentation also reports counting performance on two COCO categories - \textit{person} and \textit{zebra}.\footnote{We compare our best performing \seq model with their approach. On \emph{person}, \seq outperforms by $1.29$ $RMSE$ and $0.24$ $relRMSE$. On \emph{zebra},~\cite{RenZ16} outperforms \seq by a margin of $0.4$ $RMSE$ and $0.23$ $relRMSE$. A recent exchange with the authors suggested anomalies in their experimental setup, which may have resulted in their reported numbers being optimistic estimates of the true performance.}
\par \noindent
For both PASCAL and COCO we observe that while \ens outperforms other approaches in some cases, it does not always do so. We hypothesize that this is due to the poor performance of \glance. For detailed ablation studies on \ens see appendix.

\begin{table} \scriptsize
\setlength{\tabcolsep}{3pt}
\begin{center}
\resizebox{1\columnwidth}{!}{
\begin{tabular}{@{} l  c  c  c  c  c@{}}
\toprule
Approach & mRMSE & mRMSE-nz & m-relRMSE & m-relRMSE-nz\\
\midrule
\texttt{always-0} & 0.54 $\pm$ 0.01 & 3.03 $\pm$ 0.03 & 0.21 $\pm$ 0.00 & 1.22 $\pm$ 0.01\\
\texttt{mean} & 0.54 $\pm$ 0.00 & 2.96 $\pm$ 0.03 & 0.23 $\pm$ 0.00 & 1.17 $\pm$ 0.01\\
\texttt{always-1} & 1.12 $\pm$ 0.00 & 2.39 $\pm$ 0.03 & 1.00 $\pm$ 0.00 & 0.80 $\pm$ 0.00\\
\texttt{category-mean} & 0.52 $\pm$ 0.01 & 2.97 $\pm$ 0.03 & 0.22 $\pm$ 0.00 & 1.18 $\pm$ 0.01\\
\texttt{gt-class}   & 0.47 $\pm$ 0.00 & 2.70 $\pm$ 0.03 & 0.20 $\pm$ 0.00 & 1.08 $\pm$ 0.00\\
\detect{}  &  0.49 $\pm$ 0.00 & 2.78 $\pm$ 0.03 & 0.20 $\pm$ 0.00 & 1.13 $\pm$ 0.01\\
\midrule
\glance{}\texttt{-ft-1L} & 0.42 $\pm$ 0.00 & 2.25 $\pm$ 0.02 & 0.23 $\pm$ 0.00 & 0.91 $\pm$ 0.00\\
\glance{}\texttt{-sos-1L} & 0.44 $\pm$ 0.00 & 2.32 $\pm$ 0.03 & 0.24 $\pm$ 0.00 & 0.92 $\pm$ 0.01 \\
\sub{}\texttt{-ft-1L-$3\times3$} & 0.38 $\pm$ 0.00 & 2.08 $\pm$ 0.02 & 0.24 $\pm$ 0.00 & 0.87 $\pm$ 0.01 \\
\seq{}\texttt{-ft-$3\times3$} &\textbf{0.35 $\pm$ 0.00} &\textbf{1.96 $\pm$ 0.02} & 0.18 $\pm$ 0.00 & 0.82 $\pm$ 0.01\\
\ens{} & 0.36$\pm$ 0.00 & 1.98$\pm$ 0.02 &\textbf{0.18$\pm$ 0.00} &\textbf{0.81$\pm$ 0.01}\\
\bottomrule
\end{tabular}
}
\vspace{0pt}
\caption{\footnotesize{Counting performance on COCO Count-test set (\texttt{L} implies the number of hidden layers). \textbf{Lower} is better. \ens is a combination of \glance{}\texttt{-ft-1L}, \sub{}\texttt{-ft-1L-$3\times3$} and \seq{}\texttt{-ft-$3\times3$}. \vspace{4pt}}}
\label{table:coco}
\vspace{-20pt}
\end{center}
\end{table}

\subsection{Analysis of the Predicted Counts} 
\begin{figure}[t]
\centering     
\includegraphics[width=\columnwidth]{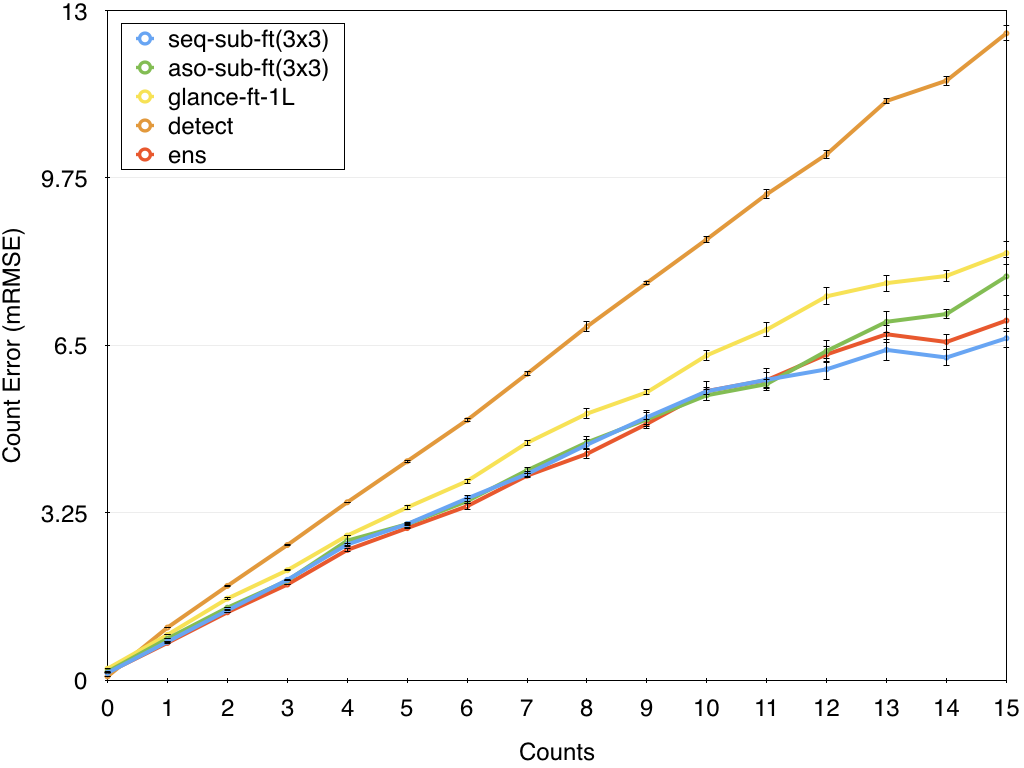}
\vspace{-20pt}
\caption{\footnotesize{We plot the $mRMSE$ (across all categories) with error bars (too small to be visible) at a count against the count (x-axis) on the Count-test split of the COCO dataset. We find that the \seq{}\texttt{-ft-$3\times3$} and \ens perform really well at higher count values whereas at lower count values the results of all the models are comparable except \detect.}}
\label{fig:a}
\vspace{-18pt}
\end{figure}
\par \noindent
\textbf{Count versus Count Error : } We analyze the performance of each of the methods at different count values on the COCO Count-test set (Fig.~\ref{fig:a}). 
We pick each count value on the x-axis and compute the $RMSE$ over all the instances at that count value. Interestingly, we find that the \emph{subitizing} approaches work really well across a range of count values. This supports our intuition that \sub and \seq are able to capture partial counts (from larger objects) as well as integer counts (from smaller objects) better which is intuitive since larger counts are likely to occur at a smaller scale. Of the two approaches, \seq works better, likely because reasoning about global context helps us capture part-like features better compared to \sub. This is quite clear when we look at the performance of \seq compared to \sub in the count range 11 to 15. For lower count values, \ens does the best (Fig.~\ref{fig:a}). We can see that for counts $>5$, \glance and \detect performances start tailing off. 
\begin{figure}[t]
\centering     
\includegraphics[width=\columnwidth]{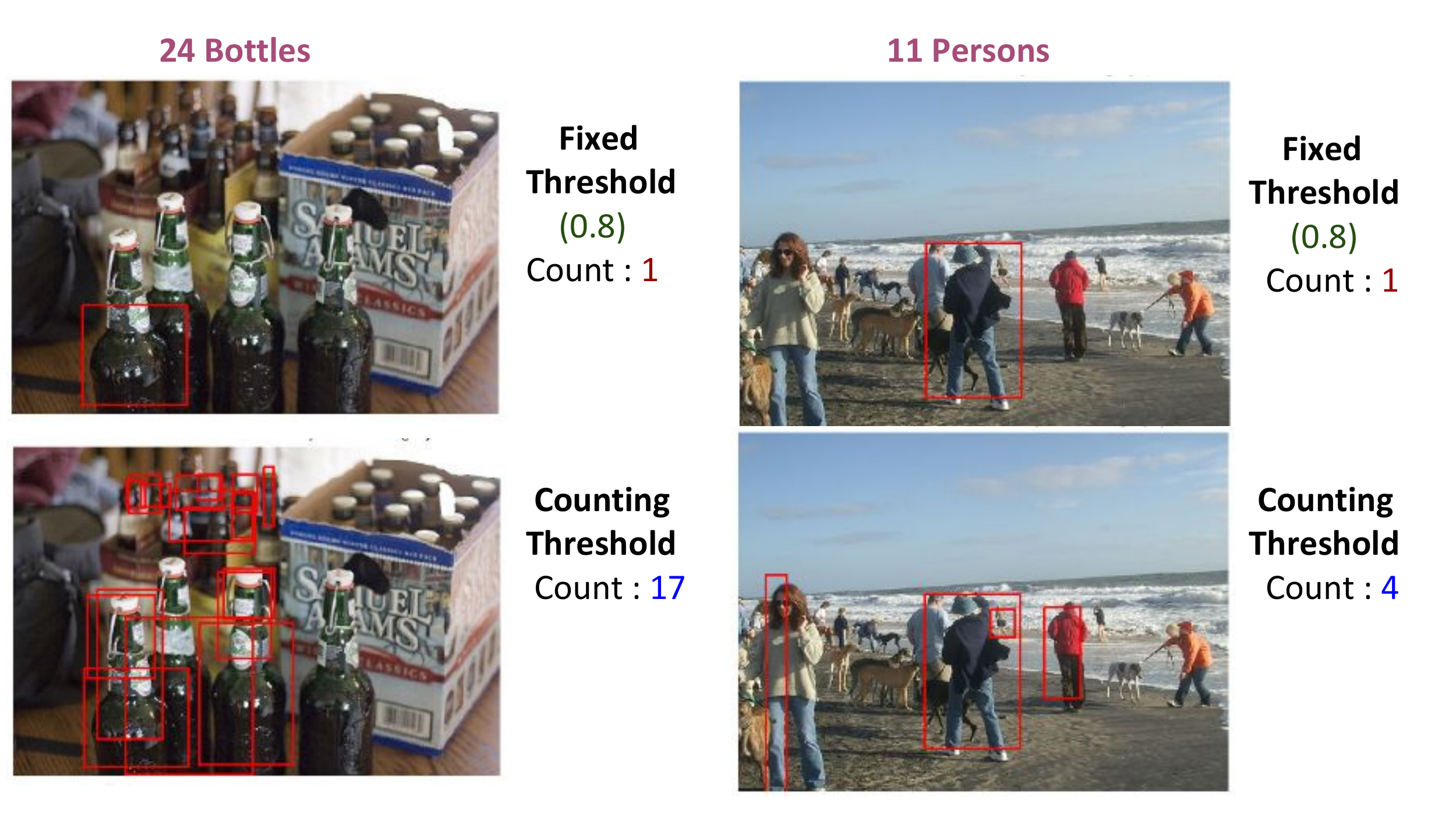}
\vspace{-15pt}
\caption{\footnotesize{We show the ground truth count (top), outputs of \detect{} with a default score threshold of 0.8 (row 1), and outputs of \detect{} with hyperparameters tuned for counting (row 2). Clearly, choosing a different threshold allows us to trade-off localization accuracy for counting accuracy (see bottle image). The method finds partial evidence for counts, even if it cannot localize the full object.}}
\label{fig:b}
\vspace{-10pt}
\end{figure}
\par \noindent
\textbf{Detection : }We tune the hyperparameters of Fast R-CNN in order to find the setting where the mean squared error is the lowest, on the Count-val splits of the datasets. We show some qualitative examples of the detection ground truth, the performance without tuning for counting (using black-box Fast R-CNN), and the performance after tuning for counting on the \pascal dataset in  Fig.~\ref{fig:b}. We use untuned Fast R-CNN at a score threshold of 0.8 and NMS threshold of 0.3, as used by Girshick \etal~\cite{Girshick2015FastR-CNN} in their demo. At this configuration, it achieves an $mRMSE$ of 0.52 on Count-test split of COCO. We find that we achieve a gain of 0.02 $mRMSE$ by tuning the hyperparameters for \detect{}.

\begin{figure}[t]
\centering
\includegraphics[width=0.8\columnwidth]{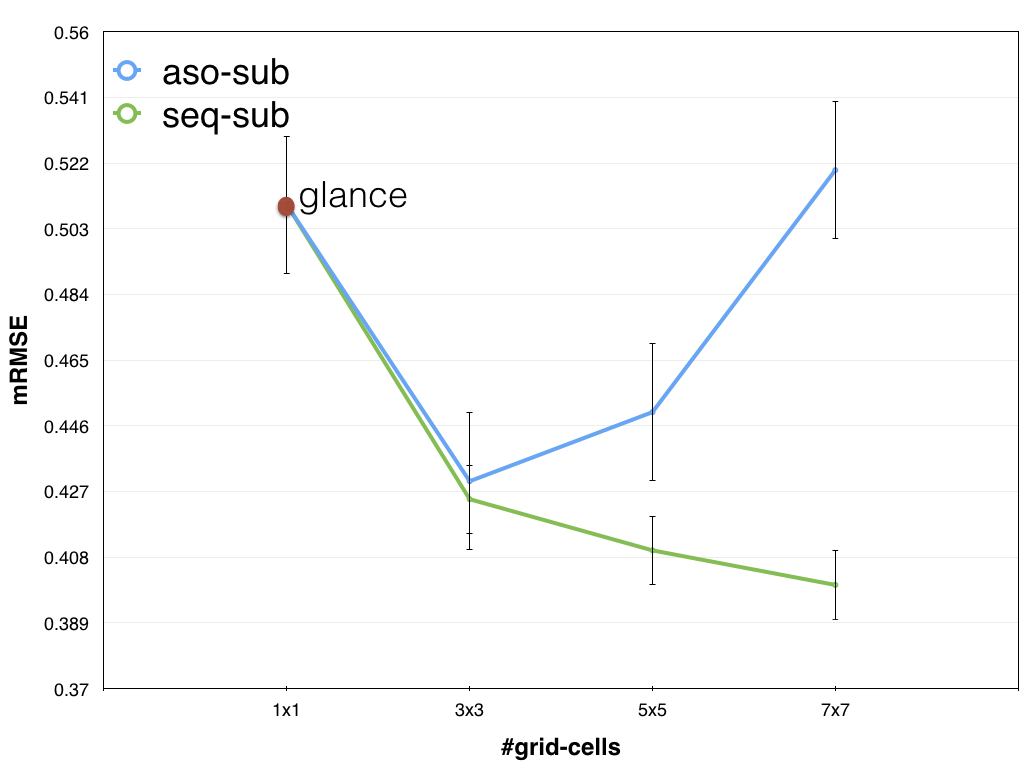}
\vspace{-4pt}
\caption{\footnotesize{We plot the $mRMSE$ across all categories (y-axis) for \sub and \seq on PASCAL Count-val set against the size of \emph{subitizing} grid cells (x-axis). As we vary the discretization we conceptually explore a
continuum between \glance{} and \detect{} approaches. We find that for \sub there exists a sweet spot (3$\times$3), where performance on counting is the best. Interestingly, for \seq the discretization sweet-spot is farther out to the right than \sub's $3\times3$.}}
\label{exp:sub-fig}
\vspace{-20pt}
\end{figure}



\par \noindent
\textbf{Subitizing : }We next analyze how different design choices in \sub{} affect performance on \pascal. We pick the best performing \sub{}\texttt{-ft-1L-$3\times3$} model and vary the grid sizes (as explained in \refsec{sec:exp_setup}). We experiment with $3\times3$, $5\times5$, and $7\times7$ grid sizes. We observe that for \sub the performance of $3\times3$ grid is the best and performance deteriorates significantly as we reach $7\times7$ grids (Fig.~\ref{exp:sub-fig}).\footnote{Going from $1\times1$ to $3\times3$, one might argue that the gain in performance in \sub is due to more (augmented) training data. However, from the diminishing performance on increasing grid size to $5\times5$ (which has even more data to train from), we hypothesize that this is not the case.} This indicates that there is indeed a sweet spot in the discretization as we interpolate between the \glance{} and \detect{} settings. However, we notice that for \seq this sweet spot lies farther out to the right.

\subsection{Counting to Improve Detection}\label{subsec:cntdet}
We now explore whether counting can help \emph{improve} detection performance (on the PASCAL dataset). Detectors are typically evaluated via the Average Precision (AP) metric, which involves a full sweep over the range of score-thresholds for the detector. While this is a useful investigative tool, in any real application (say autonomous driving), the detector must make hard decisions at some fixed threshold. This threshold could be chosen on a per-image or per-category basis. Interestingly, if we knew \emph{how many} objects of a category are present, we could simply set the threshold so that those many objects are detected similar to Zhang~\etal~\cite{zhang2015salient}. Thus, we could use per-image-per-category counts as a prior to improve detection. 

Note that since our goal is to intelligently pick a threshold for the detector, computing AP (which involves a sweep over the thresholds) is not possible. Hence, to quantify detection performance, we first assign to each detected box one ground truth box with which it has the highest overlap. Then for each ground truth box, we check if any detection box has greater than 0.5 overlap. If so, we assign a match between the ground truth and detection, and take them out of the pool of detections and ground truths. Through this procedure, we obtain a set of true positive and false positive detection outputs. With these outputs we compute the precision and recall values for the detector. Finally, we compute the F-measure as the harmonic mean of these precision and recall values, and average the F-measure values across images and categories. We call this the \textbf{mF} (mean F-measure) metric. As a baseline, we use the Fast-RCNN detector after NMS to do a sweep over the thresholds for each category on the validation set to find the threshold that maximizes F-measure for that category. We call this the \texttt{base} detector.

With a fixed per-category score threshold, the \texttt{base} detector gets a performance of 15.26\% mF. With ground truth counts to select thresholds, we get a best-case \texttt{oracle} performance of 20.17\%. Finally, we pick the outputs of \ens and \seq{}\texttt{-ft} models and use the counts from each of these to set separate thresholds. Our counting methods undercount more often than they overcount\footnote{See appendix
 for more details.}, a high count implies that the ground truth count is likely to be even higher. Thus, for counts of 0, we default to the \texttt{base} thresholds and for the other predicted counts, we use the counts to set the thresholds. With this procedure, we get a gains of 1.64\% mF and 1.74\% mF over the \texttt{base} performance using \ens and \seq{}\texttt{-ft} predictions respectively. Thus, counting can be used as a complimentary signal to aid detector performance, by intelligently picking the detector threshold in an image specific manner.

\subsection{VQA Experiment}\label{subsec:vqa_exp_paper}
We explore how well our counting approaches do on simple counting questions. Recent work~\cite{vqa,Ren2015ExploringAnswering,Malinowski2015AskImages,FukuiPYRDR16} has explored the problem of answering free-form natural language questions for images. One of the large-scale datasets in the space is the Visual Question Answering~\cite{vqa} dataset. 
We also evaluate using the COCO-QA dataset from ~\cite{Ren2015ExploringAnswering} which automatically generates questions from human captions. Around 10.28\% and 7.07\% of the questions in VQA and COCO-QA are ``how many'' questions related to counting objects. Note that both the datasets use images from the COCO~\cite{LinECCV14coco} dataset. We apply our counting models, along with some basic natural language pre-processing to answer some of these questions.

\begin{figure}[t]
\includegraphics[width=\columnwidth]{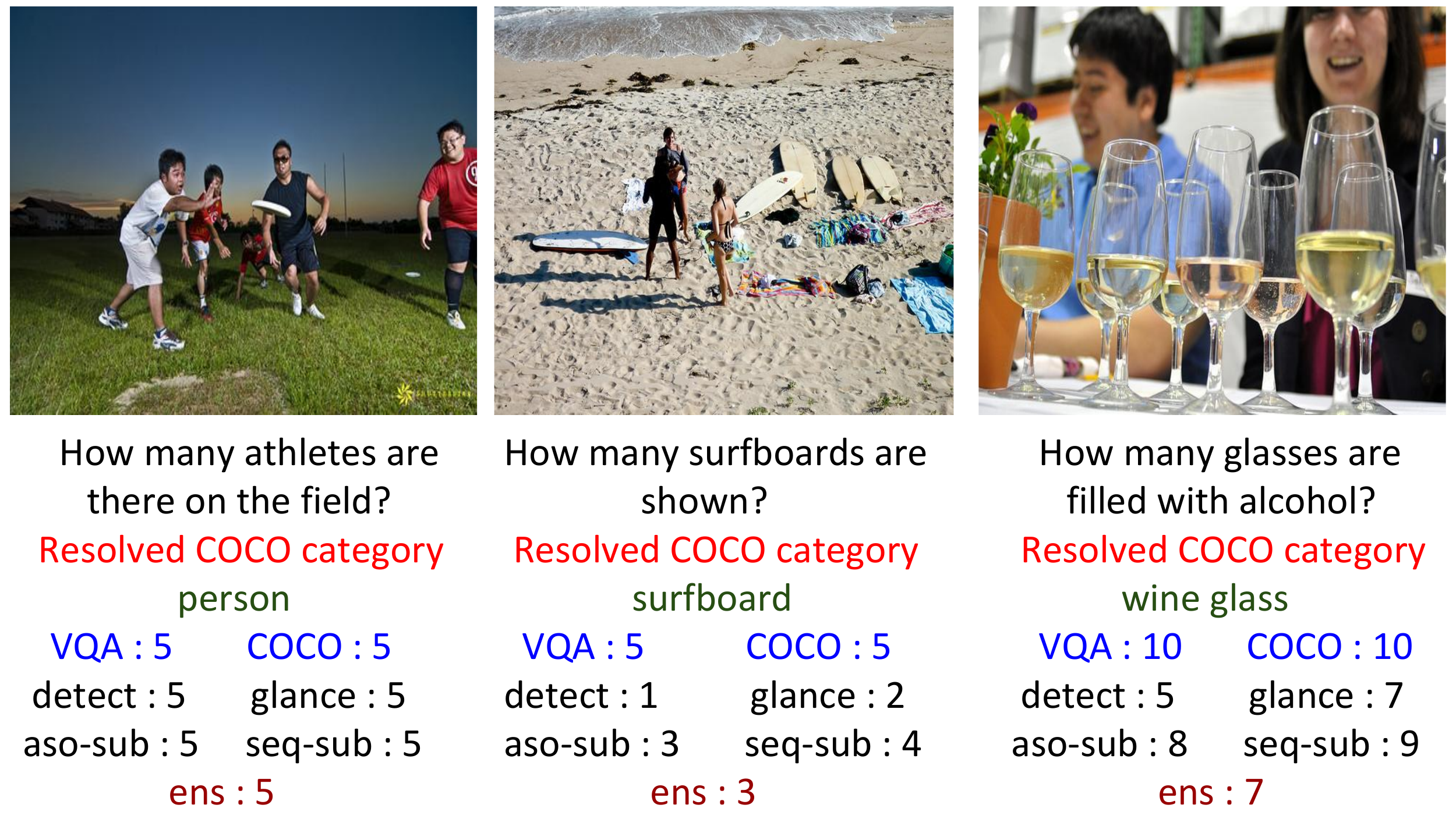}
\vspace{-15pt}
\caption{\footnotesize{Some examples from the Count-QA VQA subset. Given a question, we parse the nouns and resolve correspondence to COCO categories. The resolved ground truth category is denoted after the question. We show the VQA ground truth and COCO dataset resolved ground truth counts, followed by outputs from \detect{}, \glance, \sub{}, \seq{} and \ens{}.\vspace{4pt}}}
\label{exp:vqa}
\vspace{-15pt}
\end{figure}

\begin{table} \footnotesize
\setlength{\tabcolsep}{5.5pt}
\begin{center}
\resizebox{1\columnwidth}{!}{
\begin{tabular}{@{} l  c  c c @{}}
\toprule
Approach & mRMSE (VQA) & mRMSE (COCO-QA) \\
\midrule
\detect{}  & 2.72 $\pm$ 0.09 & 2.59 $\pm$ 0.12 \\
\glance{}\texttt{-ft-1L} & 2.19 $\pm$ 0.05 & 1.86 $\pm$ 0.12\\
\sub{}\texttt{-ft-1L-$3\times3$} & 1.94 $\pm$ 0.07 & 1.47 $\pm$ 0.04\\
\seq{}\texttt{-ft-$3\times3$} & 1.81 $\pm$ 0.09 &\textbf{1.34 $\pm$ 0.07}\\
\ens{} &\textbf{1.80 $\pm$ 0.07} & 1.40 $\pm$ 0.08\\
Deeper LSTM~\cite{Lu2015} & 2.71 $\pm$ 0.23 & N/A \\
SOTA VQA~\cite{FukuiPYRDR16} & 3.25 $\pm$ 0.94 & N/A \\
\hline
\end{tabular}
}
\caption{Performance of various methods on counting questions in the \textbf{Count-QA} splits of the VQA dataset and COCO-QA datasets respectively (\texttt{L} implies the number of hidden layers). \textbf{Lower} is better. \ens is a combination of \glance{}\texttt{-ft-1L}, \sub{}\texttt{-ft-1L-$3\times3$} and \seq{}\texttt{-ft-$3\times3$}.} 
\label{table:vqa}
\vspace{-30pt}
\end{center}
\end{table}
Given the question ``how many bottles are there in the fridge?" we need to reason about the object of interest (bottles), understand referring expressions (in the fridge) \etc Note that since these questions are free form, the category of interest might not exactly correspond to an COCO category. We tackle this ambiguity by using word2vec embeddings~\cite{Mikolov2013}. Given a free form natural language question, we extract the noun from the question and compute the closest COCO category by checking similarity of the noun with the categories in the word2vec embedding space. In case of multiple nouns, we just retain the first noun in the sentence (since how many questions typically have the subject noun first). We then run the counting method for the COCO category (see Fig~\ref{exp:vqa}). More details can be found in the supplementary. Note that parsing referring expressions is still an open research problem \cite{KazemzadehOrdonezMattenBergEMNLP14,SadovnikGC13}. Thus, we filter questions based on an ``oracle" for resolving referring expressions. This oracle is constructed by checking if the ground truth count of the COCO category we resolve using word2vec matches with the answer for the question. Evaluating only on these questions allows us to isolate errors due to inaccurate counts. We evaluate our outputs using the $RMSE$ metric. We use this procedure to compile a list of 1774 and 513 questions (\textbf{Count-QA}) from the VQA and COCO-QA datasets respectively, to evaluate on. We will publicly release our Count-QA subsets to help future work.

We report performances in Table.~\ref{table:vqa}. The trend of increasing performance is visible from \glance to \ens. We find that \seq significantly outperforms the other approaches. We also evaluate a state-of-the-art VQA model~\cite{FukuiPYRDR16} on the Count-QA VQA subset and find that even \glance does better by 
a substantial margin.\footnote{For the column corresponding to VQA, all methods are evaluated on the subset of the predictions where~\cite{Lu2015} and~\cite{FukuiPYRDR16} both produced numerical answers. For~\cite{Lu2015}, there were 11 non-numerical answers and for~\cite{FukuiPYRDR16} there were 3 (e.g., "many", "few", "lot")} 

\section{Conclusion}\label{sec:conclusion}
We study the problem of counting \emph{everyday} objects in \emph{everyday} scenes. We evaluate some baseline approaches to this problem using object detection, regression using global image features, and associative subitizing which involves regression on non-overlapping image cells. We propose sequential subtizing, a variant of the associative subitizing model which incorporates context across cells using a pair of stacked bi-directional  LSTMs. We find that our proposed models lead to improved performance on PASCAL VOC 2007 and COCO datasets. We thoroughly evaluate the relative strengths, weaknesses and biases of our approaches, providing a benchmark for future approaches on counting, and show that an ensemble of our proposed approaches peforms the best. Further, we show that counting can be used to improve object detection and present proof-of-concept experiments on answering `how many?' questions in visual question answering tasks.  
Our code and datasets will be made publicly available. 
\par \noindent
\noindent\textbf{Acknowledgements.} We are grateful to the developers of Torch~\cite{torch} for building an excellent framework. This work was funded in part by NSF CAREER awards to DB and DP, ONR YIP awards to DP and DB, ONR Grant N00014-14-1-0679 to DB, a Sloan Fellowship to DP, ARO YIP awards to DB and DP, an Allen Distinguished Investigator award to DP from the Paul G. Allen Family Foundation, Google Faculty Research Awards to DP and DB, Amazon Academic Research Awards to DP and DB, and NVIDIA GPU donations to DB. The views and conclusions contained herein are those of the authors and should not be interpreted as necessarily representing the official policies or endorsements, either expressed or implied, of the U.S. Government, or any sponsor.

{\small
\bibliographystyle{ieee}
\bibliography{egbib,rama,visw2v,Mendeley}
}

\clearpage
\section*{Appendix}\label{sec:appendix}
\setcounter{section}{0}
\normalsize
\newenvironment{packed_itemize}{
\begin{enumerate}{\labelitemi}{\leftmargin=2em}
\vspace{-6pt}
 \setlength{\itemsep}{0pt}
 \setlength{\parskip}{0pt}
 \setlength{\parsep}{0pt}
}
{\end{enumerate}}

\begin{enumerate}
\itemsep0em
\item In \refsec{sec:abl}, we report results of ablation studies conducted on the Count-val split for \glance, \sub and \seq models
\item In \refsec{sec:under_over}, we report some analyses of the count predictions generated by our models, specifically comparing object sizes with count performance and overcounting-undercounting statistics
\item In \refsec{sec:vqa_exp} we present some details of the VQA experiment performed in \refsec{subsec:vqa_exp_paper} in the main paper
\item In \refsec{sec:occ}, we show results of occlusion studies performed to identify the regions of interest in the scene while estimating the counts
\item In \refsec{sec:qual}, we present some qualitative examples of the predictions generated by our models
\end{enumerate}

\section{Ablation Studies}\label{sec:abl}
We explain the architectures for \glance, \sub and \seq in \refsec{sec:approach}. Here we report results of some ablation studies conducted on these architectures.

For \glance and \sub, we search over the following architecture space. Firstly, we vary the hidden layer sizes in the set $[250, 500, 1000, 1500, 2000]$. Secondly, we vary the number of hidden layers in the model between $1$ and $2$ (with the previously selected hidden layer size). Corresponding to these settings, we search for the best performing archiecture for \texttt{ft} (detection finetuned fc7) and \texttt{noft} (classification fc7) features extracted from PASCAL images. For \sub, in addition to this, we look for the best performing architecture across different grid sizes ($3\times3$, $5\times5$, $7\times7$). We narrow down to some design choices with $3\times3$ and them compare different grid sizes.

For \seq, we vary the number of Bi-LSTM (context aggregator) units per sequence. Subsequently, we vary the grid size from $3\times3$ to $5\times5$. We report studies on both PASCAL and COCO.

All results are reported on the Count-val splits of the concerned datasets.



\begin{figure}
\includegraphics[width=\columnwidth]{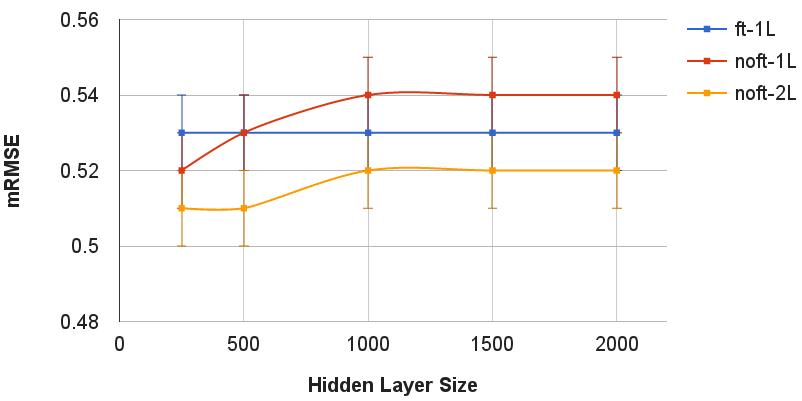}
\caption{Variations in $mRMSE$ (\textbf{Lower} is better) on the y-axis on PASCAL Count-val set with different sizes of hidden layers on the x-axis for \glance. We show performance with 1 hidden layer for \texttt{ft} (detection finetuned features) as well as \texttt{noft} (classification features). We observe that the classification features do better. We then increase the number of hidden layers to 2 for the model \texttt{noft} features and find that it does the best.}
\label{fig:pascal_glance}
\end{figure}

\textbf{\glance{} : }
We find that the performance for \texttt{ft-1L} remains more or less constant as we change the size of the hidden layers (\reffig{fig:pascal_glance}). In contrast, the \texttt{noft-1L} model does best at smaller hidden layer sizes. A two hidden layer \texttt{noft} model does better than both \texttt{1L} models. Intuitively, this makes sense since the \texttt{noft} features are better suited to global image statistics than the detection finetuned \texttt{ft} features.

\begin{figure}
\includegraphics[width=\columnwidth]{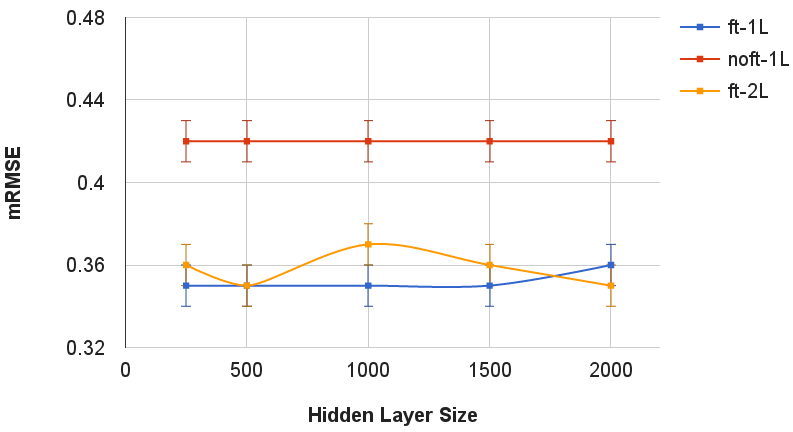}
\caption{Variations in $mRMSE$ (\textbf{Lower} is better) on the y-axis on PASCAL Count-val set with different sizes of hidden layers on the x-axis for \sub-\texttt{$3\times3$}. We show performance with 1 hidden layer for \texttt{ft} (detection finetuned features) as well as \texttt{noft} (classification features). We observe that the detection finetuned features do better. We then increase the number of hidden layers to 2 for the model \texttt{ft} features and find that it does not give a substantial performance boost over \texttt{ft}.}
\label{fig:pascal_asosub}
\end{figure}

\textbf{\sub-\texttt{$3\times3$} : }
We next contrast different design choices for \sub-\texttt{$3\times3$}. Details of how the performance changes with different grid sizes in \sub has been discussed the main paper. In particular, just like the previous section, we study the impact of hidden layer sizes and number of hidden layers, as well as the choice of features (\texttt{ft} vs \texttt{noft}) for the \sub-\texttt{$3\times3$} model (\reffig{fig:pascal_asosub}). We find that the detection finetuned \texttt{ft} features do much better than the classification features \texttt{noft} for \sub. This is likely because the \texttt{ft} features are better adapted to statistics of local image regions than the \texttt{noft} image classification features. We also find that increasing the number of hidden layers does not improve performance over using a single hidden layer, unlike \glance. 

\begin{figure}
\includegraphics[width=\columnwidth]{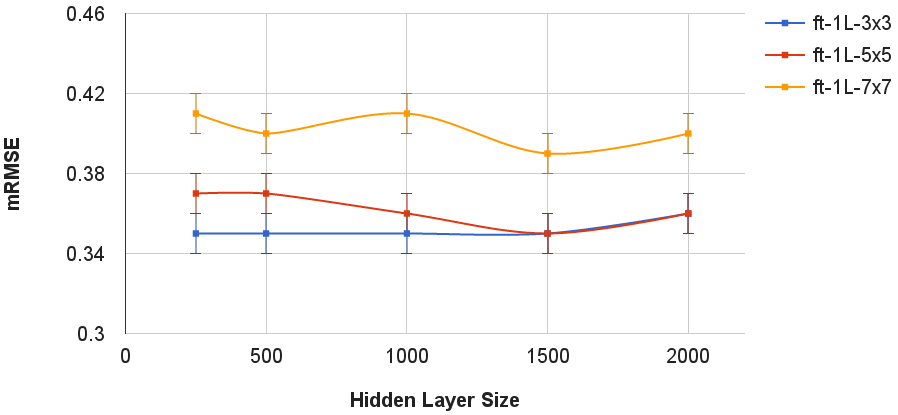}
\caption{Variations in $mRMSE$ (\textbf{Lower} is better) on the y-axis on PASCAL Count-val set with different sizes of hidden layers on the x-axis for \sub. We compare the grid sizes $3\times3$, $5\times5$, and $7\times7$, and find that the $3\times3$ setting performs the best.}
\label{fig:pascal_asosubfine}
\end{figure}

\textbf{\sub{} : }
We next compare how the performance of \sub varies as we change the size of the grids. We pick the best performing \sub-\texttt{$3\times3$} features (\texttt{ft}) and number of hidden layers - 1. We then vary the size of the hidden layer and compare the performance of $3\times3$, $5\times5$, and $7\times7$ \sub approaches (\reffig{fig:pascal_asosubfine}). We find that $3\times3$ and $5\times5$ models do much better than the $7\times7$ model. The performance of $3\times3$ is slightly better than the $5\times5$ model. A similar comparison on the Count-test set can be found in the main paper.

\begin{figure}
\includegraphics[width=\columnwidth]{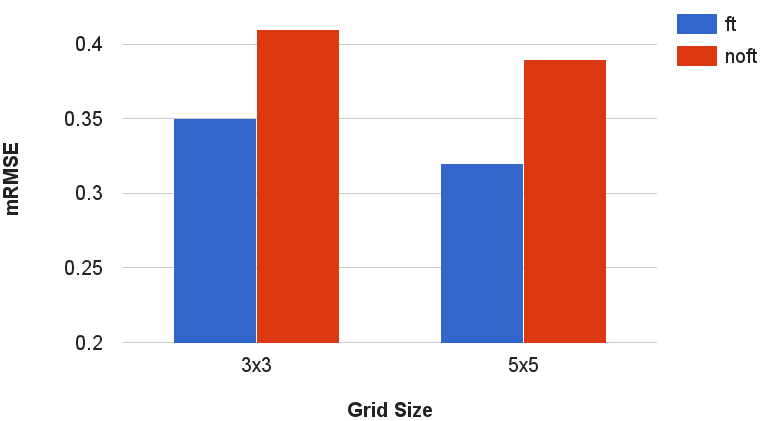}
\caption{$mRMSE$ (\textbf{Lower} is better) on the y-axis on the PASCAL Count-val set as a result of varying the grid size on the x-axis for \seq. We experiment with both \texttt{ft} and \texttt{noft} features. On going from $3\times3$ to $5\times5$, there is an improvement in performance. Note that models using \texttt{noft} features perform worse than models using \texttt{ft} features.}
\label{fig:pascal_seq}
\end{figure}

\begin{figure}
\includegraphics[width=\columnwidth]{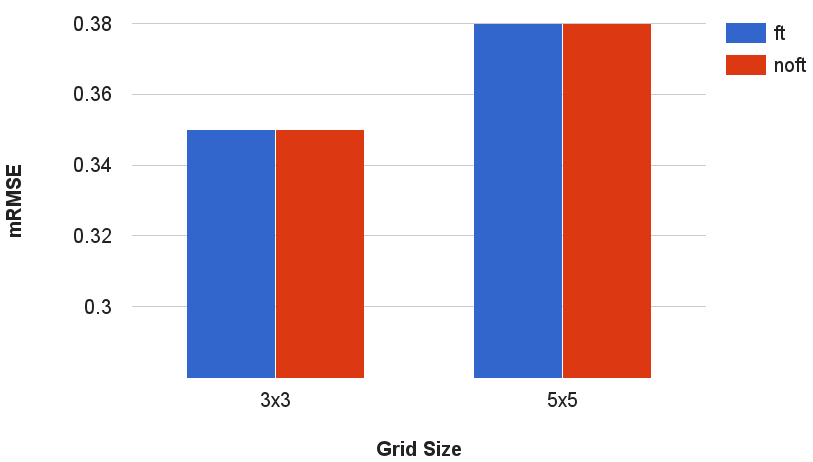}
\caption{$mRMSE$ (\textbf{Lower} is better) on the y-axis on the COCO Count-val set as a result of varying the grid size on the x-axis for \seq. We experiment with both \texttt{ft} and \texttt{noft} features. On going from $3\times3$ to $5\times5$, there is a decrease in performance.}
\label{fig:coco_seq}
\end{figure}

\textbf{\seq{} : }
In \reffig{fig:pascal_seq} and \reffig{fig:coco_seq}, we compare the effect of changing the grid size from $3\times3$ to $5\times5$ for the \seq models. We use both \texttt{ft} and \texttt{noft} features extracted from PASCAL and COCO images. On PASCAL (\reffig{fig:pascal_seq}), we observe that increasing the grid size has a slight improvement in performance for both \texttt{ft} and \texttt{noft} features unlike COCO (\reffig{fig:coco_seq}) where going from $3\times3$ to $5\times5$ there is a drop in performance for both \texttt{ft} and \texttt{noft} features. We should also note that in general \texttt{ft} features perform better than \texttt{noft} features similar to \sub.

We also varied the number of Bi-LSTM (context aggregator) units from $2$ to $1$ per sequence in the \seq architectures for a grid size of $3\times3$. We observe that for \texttt{ft} features, change in the number of Bi-LSTM units does not make a difference on both PASCAL and COCO. However, for \texttt{noft} features, going from $2$ to $1$ leads to a drop of $0.01$ $mRMSE$ on COCO and PASCAL.



\section{Count Analysis}\label{sec:under_over}
\begin{figure}
\includegraphics[width=\columnwidth]{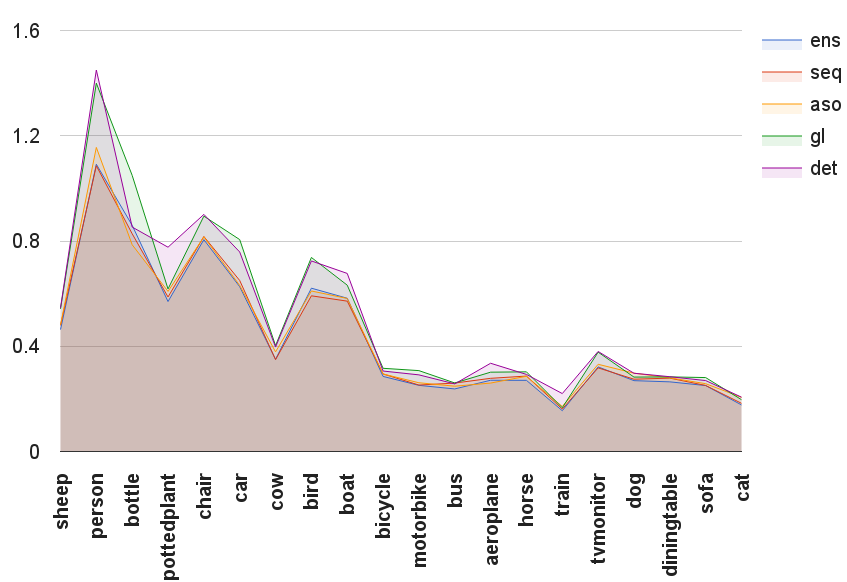}
\caption{We visualize $RMSE$ (\textbf{Lower} is better) on the y-axis for \glance (gl), \detect (det), \sub (aso), \seq (seq) and \ens (ens) across categories of various sizes on PASCAL. On the x-axis we order categories in increasing order of object size from left to right. As the object size increases, all the methods start performing competitively. We find that \seq, \sub and \ens perform consistently well for a wide range of category sizes.}
\label{fig:size_pascal}
\end{figure}

\begin{figure*}
\includegraphics[width=\textwidth]{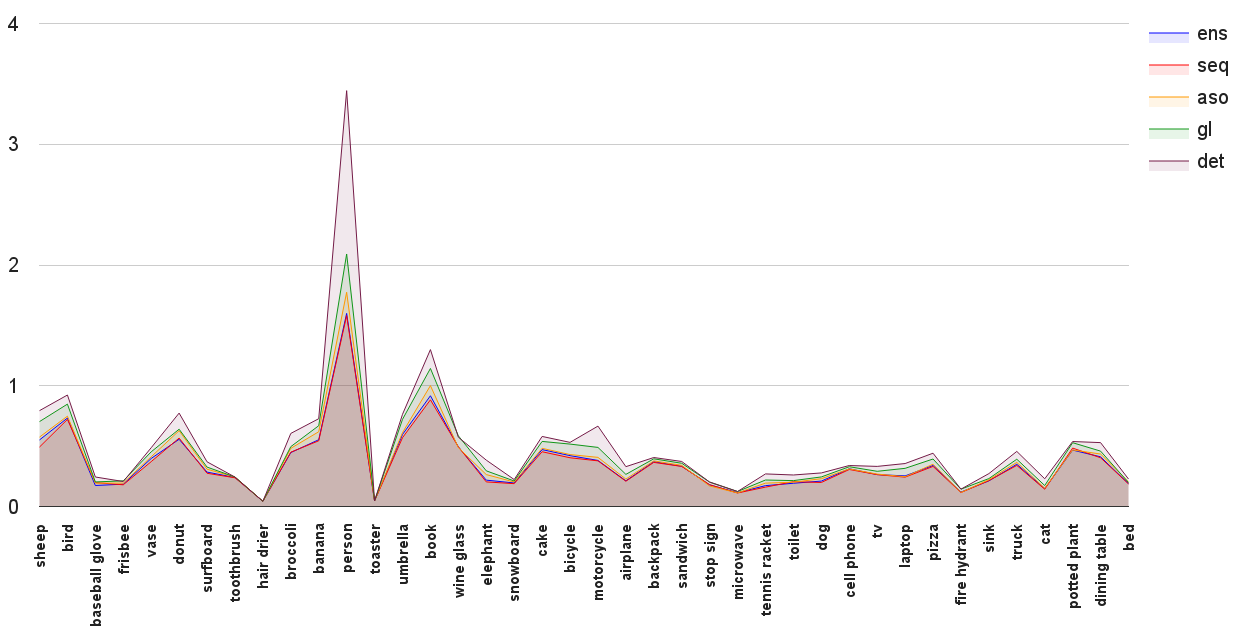}
\caption{We visualize $RMSE$ (\textbf{Lower} is better) on the y-axis for \glance (gl), \detect (det), \sub (aso), \seq (seq) and \ens (ens) across categories of various sizes on COCO. On the x-axis we order categories in increasing order of object size from left to right. We find that \seq, \sub and \ens perform consistently well for a wide range of category sizes.}
\label{fig:size_coco}
\end{figure*}

\textbf{Size versus Count Error : }
We compare \seq, \glance, \sub and \detect their performance for object categories of various sizes on PASCAL (\reffig{fig:size_pascal}) and COCO (\reffig{fig:size_coco}). To get the object size, we sum the number of pixels occupied by an object across images where the object occurs in the Count-val set and divide this number by the average number of (non-zero) instances of the object. This gives us an estimate of the expected size occupied per instance of an object. We show a sorting of smaller to larger categories on the x-axis in \reffig{fig:size_pascal} and \reffig{fig:size_coco}. 
We find that \ens, \seq and \sub perform consistently well across the spectrum of object sizes on both PASCAL and COCO. On PASCAL, as the object size increases the error keeps on reducing. This trend is not consistent over the entire spectrum of sizes for COCO. Another interesting thing to observe is that as the object size increases, the methods start performing competitively. This also indicates that \sub and \seq are able to capture partial ground truth counts well, since the counts for larger categories will necessarily be partial.


\textbf{Undercounting versus Overcounting : }
\begin{figure*}
\includegraphics[width=0.97\textwidth]{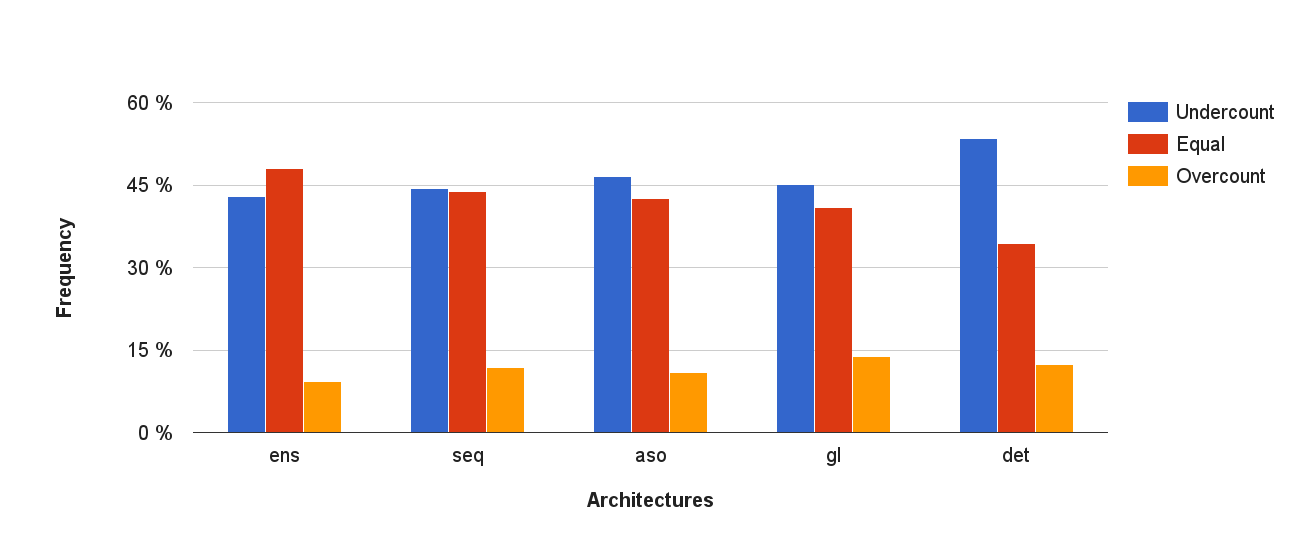}
\caption{We plot the percentage number of times \ens (ens), \seq (seq), \sub(aso), \glance (gl) and \detect (det) undercount, overcount and predict the ground truth count on PASCAL Count-test split. Going from \detect to \ens there is a steady increase in the number of times we get the count right.}
\label{fig:pascal_freq}
\end{figure*}
\begin{figure*}
\includegraphics[width=0.97\textwidth]{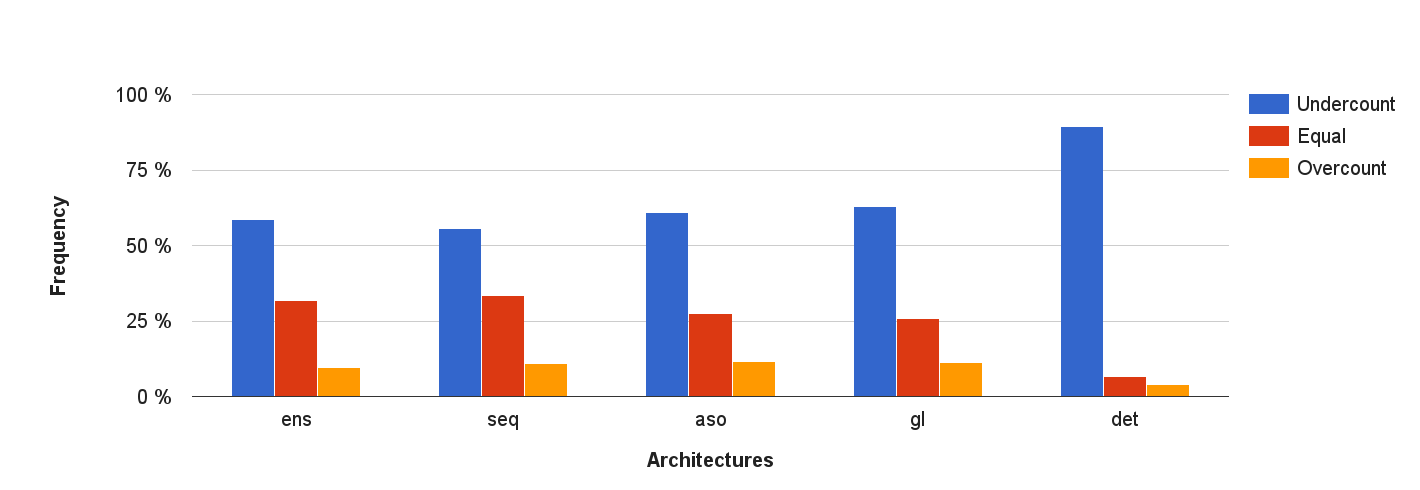}
\caption{We plot the percentage number of times \ens (ens), \seq (seq), \sub(aso), \glance (gl) and \detect (det) undercount, overcount and predict the ground truth count on COCO Count-test split. Although, going from \detect to \ens there is a steady increase in the number of times we get the count right but we undercount a lot more than getting the count right.}
\label{fig:coco_freq}
\end{figure*}
We study whether the models proposed in the paper undercount or overcount. Specifically, we report the number of times the approaches overcount, undercount or predict the ground truth count on PASCAL (\reffig{fig:pascal_freq}) and COCO (\reffig{fig:coco_freq}). To do this, we first filter out all instances where the ground truth count is 0 (since we cannot undercount 0). We then check if the predicted count is greater than the ground truth (overcounting) or lesser (undercounting) or equal to the ground truth (equal). We perform this analysis on the Count-test split. 

On PASCAL (\reffig{fig:pascal_freq}), we can observe that there is a clear increase in the number of times we get the count from \detect to \ens. The models, in general undercount more often than they overcount. As we go from \detect to \ens, the improvement in performance can be accounted to the increase of the frequency of equal versus undercount. Interestingly, for ens we get the count right more number of times as opposed to undercounting the ground truth. The number of times we overcount more or less stays the same.

On COCO (\reffig{fig:coco_freq}), we observe that although there is an increase in the number of times we get the count right as we go from \detect to \ens, the frequency of equal is much lower than the frequency of undercounting for all the models. This is understandable as COCO has more number of categories and objects of different categories have lesser chances of being in the same image.


\textbf{Ensemble : }
We study different combinations of the predictions for constructing the ensemble on the Count-test set.

On PASCAL, when we compose an ensemble of \seq and \sub, we get a $mRMSE$ of 0.427 as opposed to a $mRMSE$ of 0.438 with \seq and \glance. One can think of combinining global and local context by taking an ensemble of \glance and \sub. However, we observe that such an ensemble underperforms when compared to \seq by 0.02 $mRMSE$. We also consider including the \detect baseline in the ensemble. We see that an ensemble of \detect, \glance, \sub and \seq gives an error of 0.43 $mRMSE$ as opposed to an ensemble of \glance, \sub and \seq which gives $mRMSE$ 0.42. Thus \detect, when included in the ensemble hurts the counting performance.

On COCO, when we compose an ensemble of \seq and \sub, we get a $mRMSE$ of 0.351 as opposed to a $mRMSE$ of 0.363 with \seq and \glance. We observe that an ensemble of \glance and \sub underperforms when compared to \seq by 0.02 $mRMSE$. When \detect is included we see that an ensemble of \detect, \glance, \sub and \seq gives an error of 0.38 $mRMSE$ as opposed to an ensemble of \glance, \sub and \seq which gives $mRMSE$ 0.36. Just like on PASCAL, \detect when included in the ensemble hurts the counting performance.

\section{Occlusion Studies}\label{sec:occ}
\begin{figure*}
\includegraphics[width=0.9\textwidth]{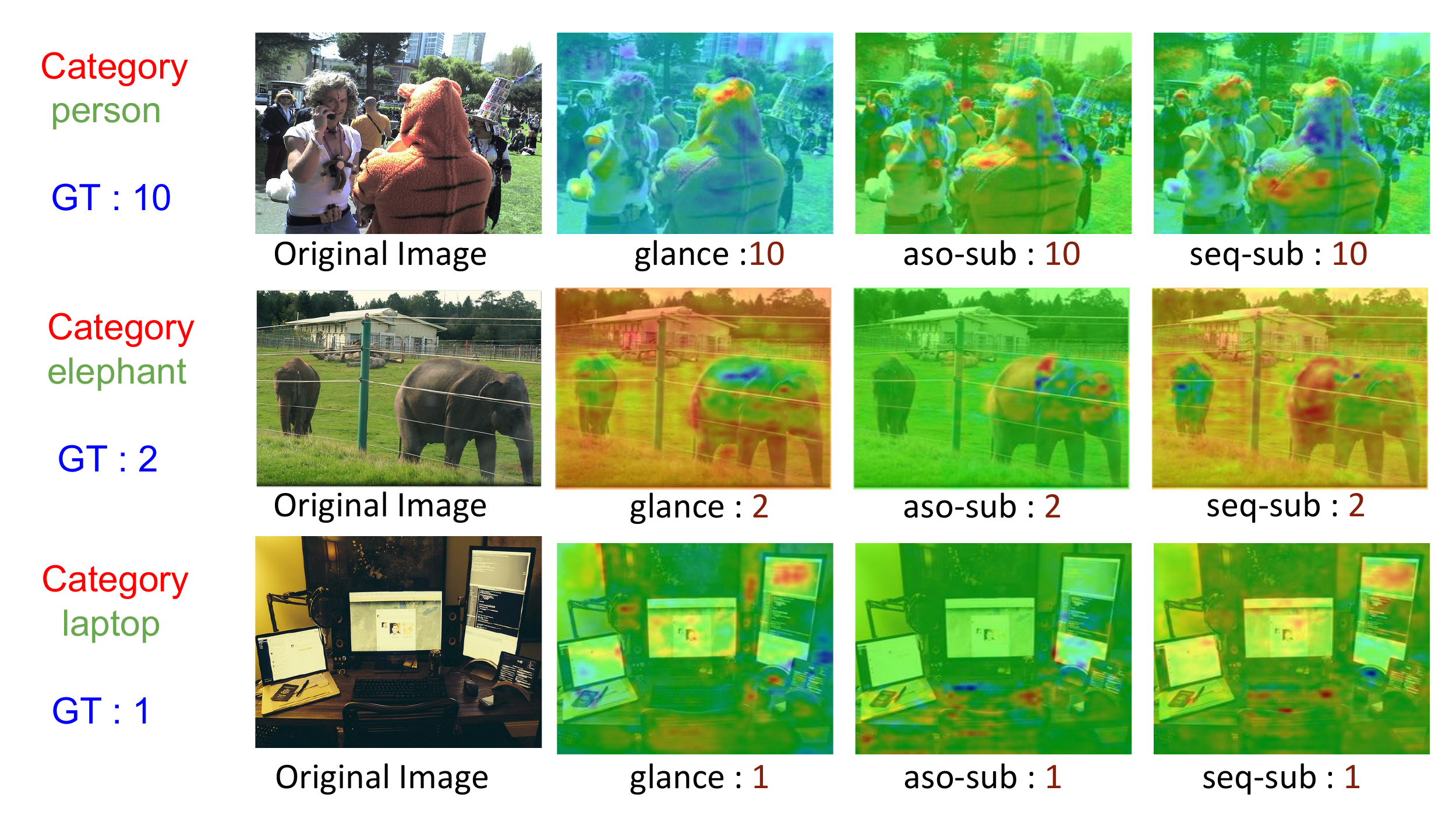}
\caption{Occlusion maps obtained on COCO Count-test images for \glance, \seq and \sub by moving a mask of size $4\times4$ over the images before doing a forward pass through each of the models. Notice how \glance{} and \seq{} approaches tend to look at similar regions for lower counts, while for higher counts \seq{} and \sub{} tend to be more similar.}
\label{fig:occ}
\vspace{-10pt}
\end{figure*}

In \reffig{fig:occ}, we perform occlusion studies to understand where \glance, \sub and \seq look in the image while estimating the counts of different objects. 

For this analysis, we pick images with a spread in counts from 10 (top row) to 1 in the middle row. For each count, we identified images where all three approaches agreed on the counts so that we could analyze where each method ``looks'' in order to derive the corresponding counts. We pick images from the COCO Count-test split where the predicted counts for \glance, \sub and \seq are equal. The \sub and \seq models are trained on $3\times3$ discretization of the images. We move $4\times4$ sized masks across the image with a non-overlapping stride to get occlusion maps. It is interesting to observe that for the image with a large ground truth count, the occlusion maps from \seq{} are very similar to those from \sub. This confirms our intuition that for larger counts, one needs access to local texture like patterns to accumulate count densities across the image. For smaller counts (rows 2 and 3), we notice that the maps from \glance and \seq are more similar, indicating that global cues such as the number of parts appearing in the image (say the number of tails of elephants), potentially captured by the distributed CNN representation are sufficient for counting. Thus, this experiment confirms our intuition that \seq captures the best of both the \glance{} and \sub{} approaches, providing us a way to ``interpolate'' between these approaches based on the counts.

\section{VQA Experiment}\label{sec:vqa_exp}
We next elaborate on more details of the VQA experiment described in \refsec{subsec:vqa_exp_paper}. More specifically we discuss how we pre-process ground truth to make it numeric and give details of how we solve correspondence between a noun in a question to counts of coco categories.

As reported in the paper, we use the  VQA dataset~\cite{vqa} and COCO-QA~\cite{Ren2015ExploringAnswering} datasets for our counting experiments. We extract the \emph{how many?} type questions, which have numerical answers. This includes both integers (VQA) and numbers written in the form of text (COCO-QA). We parse the latter into corresponding numbers on the COCO-QA dataset. That is \emph{five} is mapped to $5$. From the selected questions, we extract the Nouns (singular, plural, and proper), and convert them to their singular form using the Stanford Natural Language Parser (NLTK)~\cite{nltk}.

We train word2vec word embeddings on Wikipedia\footnote{https://www.wikipedia.org/} and use cosine similarities in the embedding space as word similarity. Using these we find the COCO category or COCO super-category that matches the most with the extracted nouns. These super-category annotations are available as part of the COCO dataset. We run our models for the COCO category, and consider it the answer. For the extracted nouns, if the best match is with a COCO super category, we sum the counts obtained by our counting methods for each of the COCO sub-categories belonging to the particular super-category. For example, if the resolved noun is \emph{animal}, we sum the counts for \emph{horse}, \emph{giraffe}, \emph{cat}, \emph{dog}, \emph{zebra}, \emph{sheep}, \emph{cow}, \emph{elephant}, \emph{bear}, and \emph{bird} and use the output as our predicted count.
\section{Qualitative Results}\label{sec:qual}
\begin{figure*}
\includegraphics[width=\textwidth]{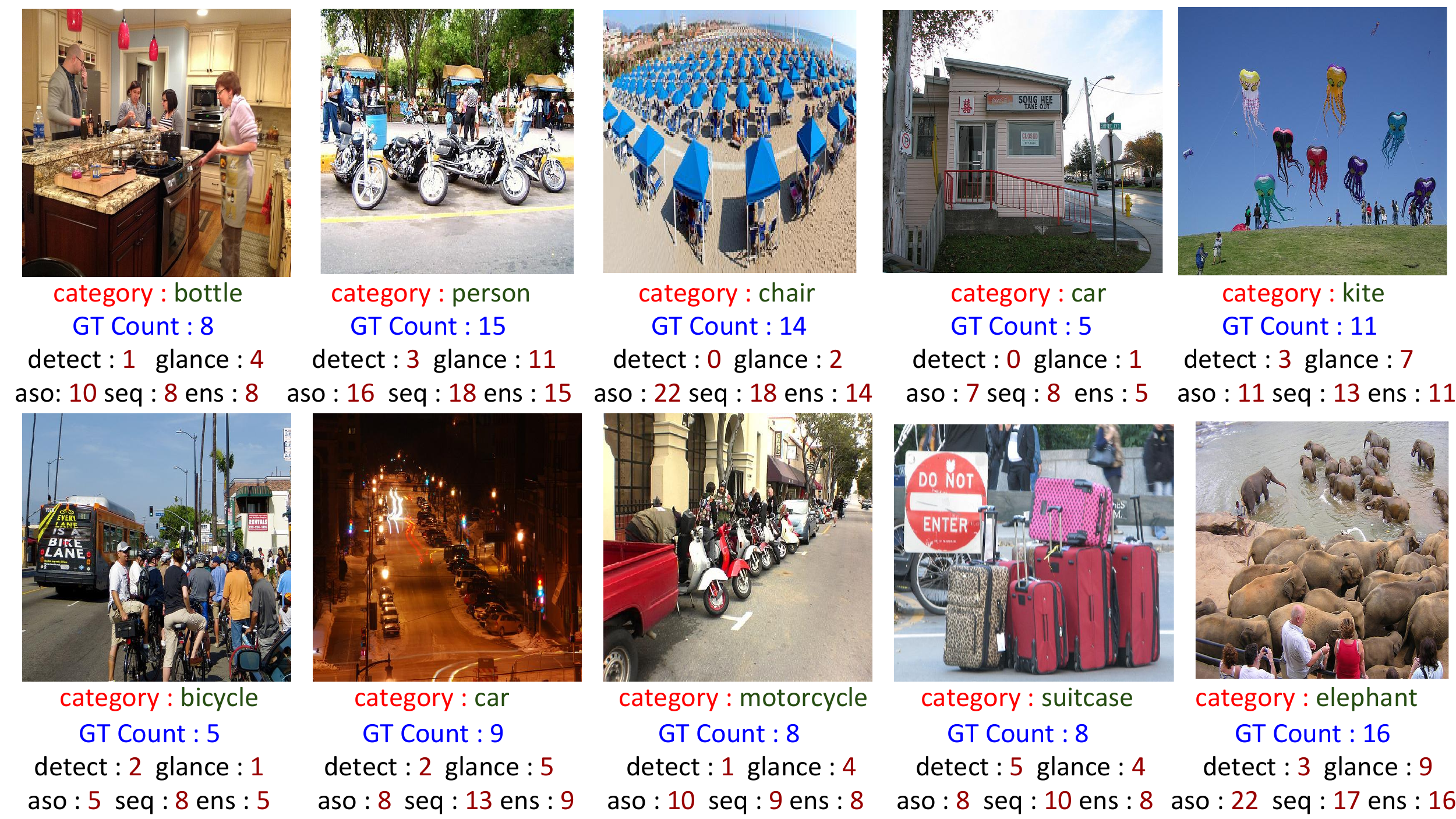}
\caption{Some qualitative examples of the predictions given by our models where \ens performs well. We can see that in cases where the objects are significantly occluded, \detect has very poor performance compared to \seq or \ens.} 
\label{fig:qual}
\end{figure*}
We finally show some qualitative examples of our predictions on COCO images in \reffig{fig:qual} where \ens performs best. We can observe that whenever the objects present in the image are sufficiently salient, \seq and \sub do a sufficiently better job in estimating the count of objects as compared to \glance. This is because as \seq, and \sub have to estimate partial counts at cell levels unlike \glance which has to regress to the count of the entire image. For some cases when the objects present are highly occluded, we see that \seq  and \sub do a much better job at estimating the count. In summary, we find that \ens as a combination of \glance, \sub and \seq gets the count right most number of times. 






\end{document}